\documentclass[journal]{IEEEtran}

\usepackage{pgfplots}
\usepackage{amsmath}
\usepackage{relsize}
\usepackage{booktabs}
\usepackage{hyperref}
\usepackage[ruled,vlined]{algorithm2e}

\usepackage{enumitem}

\usepackage{textcomp}

\pgfplotsset{compat=newest}

\newcommand\CC{C\nolinebreak[4]\hspace{-.05em}\raisebox{.4ex}{\relsize{-3}{\textbf{++}}}}

\newcommand{\rpm}{\raisebox{.2ex}{\(\scriptstyle\pm\)}}

\newcommand{\func}[1]{\operatorname{#1}}

\usetikzlibrary{fit}
\usetikzlibrary{pgfplots.statistics}

\begin{document}

\title{Real-Time Optical Flow for Vehicular Perception with Low- and High-Resolution Event Cameras}

\author{Vincent~Brebion, Julien~Moreau, and~Franck~Davoine% <-this % stops a space
\thanks{Manuscript received...}% <-this % stops a space
\thanks{V. Brebion, J. Moreau, and F. Davoine are with Universit\'e de technologie de Compi\`egne (UTC), CNRS, Heudiasyc (Heuristics and Diagnosis of Complex Systems), CS 60 319 - 60 203 Compi\`egne Cedex, France (e-mail: \textit{firstname}.\textit{lastname}@hds.utc.fr)}% <-this % stops a space
\thanks{This work was supported in part by the Hauts-de-France Region, and by the SIVALab joint lab (Renault - UTC - CNRS).}}

\markboth{IEEE TRANSACTIONS ON INTELLIGENT TRANSPORTATION SYSTEMS}%
{Brebion \MakeLowercase{\textit{et al.}}: Real-Time Optical Flow for Low- and High-Resolution Event Cameras}


\maketitle

\begin{abstract}
Event cameras capture changes of illumination in the observed scene rather than accumulating light to create images. Thus, they allow for applications under high-speed motion and complex lighting conditions, where traditional frame-based sensors show their limits with blur and over- or under-exposed pixels. Thanks to these unique properties, they represent nowadays an highly attractive sensor for ITS-related applications. Event-based optical flow (EBOF) has been studied following the rise in popularity of these neuromorphic cameras. The recent arrival of high-definition neuromorphic sensors, however, challenges the existing approaches, because of the increased resolution of the events pixel array and a much higher throughput. As an answer to these points, we propose an optimized framework for computing optical flow in real-time with both low- and high-resolution event cameras. We formulate a novel dense representation for the sparse events flow, in the form of the ``inverse exponential distance surface''. It serves as an interim frame, designed for the use of proven, state-of-the-art frame-based optical flow computation methods. We evaluate our approach on both low- and high-resolution driving sequences, and show that it often achieves better results than the current state of the art, while also reaching higher frame rates, 250Hz at 346\texttimes{}260 pixels and 77Hz at 1280\texttimes{}720 pixels.
\end{abstract}


% \fi
\IEEEpeerreviewmaketitle{}

\section{Introduction}
\IEEEPARstart{O}{ver} the last decade, neuromorphic cameras have risen in popularity, due to their unmatched qualities: low latency, high dynamic range, no motion blur, and low energy consumption~\cite{Gallego2020EventbasedVA}. Thanks to their asynchronous response and their low latency, event sensors are by nature well suited for dynamic scenes analysis, including optical flow.

Optical flow depicts the per-pixel displacement in an image after a short period (e.g., between consecutive frames). It is usually computed using the brightness constancy constraint \cite{Horn1981DeterminingOF}: pixels intensities remain constant over short durations. Optical flow is a key enabler for many major applications, such as object detection and tracking~\cite{Braillon2006RealtimeMO, Huang2018OpticalFB}, motion estimation~\cite{Hossen2016ASS}, visual odometry~\cite{Chuanqi2017MonocularVO, Tang2020FastSV}, and image segmentation~\cite{Galic2000SpatiotemporalIS}.

However, there is no direct translation for frame-based algorithms to event cameras. The sparse and asynchronous nature of their output constitutes a major paradigm shift.

Considering this, several event-based optical flow (henceforth EBOF) methods have been proposed. Former approaches have defined EBOF as a spatiotemporal point cloud matching problem, solved by plane-fitting algorithms~\cite{Benosman2014EventBasedVF,Akolkar2020SeeBY}. Others have made the choice to accumulate events during short time windows, to create dense image-like representations. They transfer the event-based problem into a frame-based one~\cite{Zhu2017EventbasedFT,Almatrafi2020DistanceSF}. The past few years have also seen the rise of neural networks to solve the EBOF problem~\cite{Zhu2018EVFlowNetSO,ParedesValls2021BackTE}.

Presented work is motivated by the arrival of new, high-resolution event sensors. While low-resolution sensors have been the reference for the past decade, high-resolution neuromorphic sensors now start to be produced~\cite{Finateu2020510A1}. They offer increased visual details, essential for advanced driving applications. Applying the aforementioned EBOF approaches to these new cameras, however, reveals a common flaw: they were all designed with low-resolution event cameras in mind. For high-resolution sensors, they output degraded or incorrect optical flow results, and hardly handle their much higher throughput, resulting in long computation times. This last issue forbids the use of these methods in the real world for intelligent transportation system applications.

As an answer to these issues, we propose a novel optimized framework for computing \emph{real-time}\footnote{Considering a car at 120Km/h, if we tolerate to drive a distance of 1 meter to achieve perception analysis, it means a maximum latency of 30ms.} EBOF for both low- and high-resolution event cameras. Our approach was originally inspired by the work of Almatrafi \textit{et al.}~\cite{Almatrafi2020DistanceSF}, who introduce a simple and efficient way of densifying events in successive frame-based representations. In this article, we propose key contributions for making the method faster, more robust, and compatible with high-definition sensors:
\begin{itemize}
    \item a specific pipeline-based architecture, for computing real-time optical flow using the events from low- or high-resolution neuromorphic sensors;
    \item the formulation of a novel dense ``inverse exponential distance surface'', that acts as the frame-based representation computed from the events, able to feed any image-based optical flow method;
    \item a coherent choice of algorithms and methods together for all the steps up to the fast frame-based state-of-the-art optical flow (with temporal smoothing to fit well with potentially noisy input events);
    \item we finally build and share a complementary high-definition event-based dataset of indoor sequences with high-speed movements, used as part of our evaluation.
\end{itemize}

Videos accompanying this article, showing results for both low- and high-resolution data, are available at \url{https://youtube.com/playlist?list=PLLL0eWAd6OXBRXli-tB1NREdhBElAxisD}.

In complement, our dataset and all the source code linked to this article are available at \url{https://github.com/vbrebion/rt_of_low_high_res_event_cameras}.

%\IEEEpubidadjcol

\section{Related Work}

\subsection{Optical Flow for Vehicular Perception}
Optical flow is of great interest in the field of perception for intelligent vehicles, where multiple dynamics are omnipresent. In~\cite{Braillon2006RealtimeMO}, Braillon \textit{et al.} compare optical flow for the ground plane seen from a moving vehicle with theorical optical flow in order to detect obstacles. A similar obstacle detection problem is explored in \cite{Huang2018OpticalFB}, where the authors use their optical flow results to dynamically determine a model of the background motion, and extract obstacles that appear as outliers. Visual odometry using optical flow has also been explored in \cite{Chuanqi2017MonocularVO} and \cite{Tang2020FastSV}, where the authors of both articles use an optical flow tracking and feature matching method for estimating the ego motion of the vehicle. An extension of optical flow to 3D, known as ``scene flow'', has also caught up the attention recently, and approaches such as \cite{Jund2021ScalableSF} aim at improving the detection of independently moving objects. Driving benchmarks such as KITTI~\cite{Geiger2012AreWR}, that includes both optical flow and scene flow, have also constituted major milestones for improving the state of the art.

\subsection{Frame-Based Optical Flow}
The problem of optical flow with traditional frame-based sensors has been deeply explored. Historical approaches relied on region-based matching techniques~\cite{Lucas1981AnII}, or on the use of spatiotemporal derivatives of the input images~\cite{Horn1981DeterminingOF}. More recent works have proposed extensions to these approaches, by defining pyramidal-based frameworks~\cite{Bouguet1999PyramidalIO, Meinhardt2013HornSchunckOF, Adarve2016AFF}, or by introducing regularization terms to add robustness~\cite{Black1996TheRE, Sun2013AQA}. The past few years have also seen the rise of neural networks, and their capability of learning from the data to generalize even in presence of noise and inconsistencies. Now, they surpass traditional handcrafted methods for optical flow and currently stand as the state of the art~\cite{Dosovitskiy2015FlowNetLO, Meister2018UnFlowUL, Teed2020RAFTRA}.

\subsection{Event-Based Optical Flow (EBOF)}\label{sec:related_work:eb_flow}
Two main approaches can be distinguished for EBOF.

On one hand, some authors use the events and all their properties, without accumulation into frame-based representations. Such a frameless approach is proposed in~\cite{Benosman2014EventBasedVF}, where is employed a plane-fitting method on short temporal windows of events, to determine their motion in the visual scene. Other works~\cite{Gallego2018AUC,Stoffregen2018SimultaneousOF,Liu2020GloballyOC} propose contrast maximization schemes as proxies for computing optical flow, by evaluating the sharpness of motion-compensated images of accumulated events. More recently, authors such as~\cite{ParedesValls2020UnsupervisedLO,ParedesValls2021SelfSupervisedLO} exploit spiking neural networks for a full bio-inspired EBOF estimation.

On the other hand, due to the great advances on optical flow estimation using traditional frames, other authors have proposed to build image-like representations from the event flow, to use them as input for these state-of-the-art methods. In~\cite{Almatrafi2020DistanceSF}, Almatrafi \textit{et al.}\ accumulate events in short temporal windows using the distance transform, to create stable dense images designed to be used with any frame-based optical flow method. In~\cite{Zhu2017EventbasedFT}, Zhu \textit{et al.}\ also create images of accumulated events, but propose to compute a sparse optical flow by extracting visual features through the use of the Harris corner detector~\cite{Harris1988ACC}, and to track them using an expectation maximization algorithm. An alternative approach is proposed in~\cite{Nagata2021OpticalFE}, where the authors describe a surface matching approach on short time-shifted images of accumulated events (time surfaces), to evaluate their displacement. Recent works have also adapted proven neural network architectures for a use with images of events; \cite{Zhu2018EVFlowNetSO, Lee2020SpikeFlowNetEO} proposed FlowNet-inspired~\cite{Dosovitskiy2015FlowNetLO} networks for inferring EBOF, while authors of \cite{Gehrig2021DenseOF} proposed a RAFT-inspired~\cite{Teed2020RAFTRA} one. Finally, Paredes-Vall\'es \textit{et al.}~\cite{ParedesValls2021BackTE} designed a light and real-time network, called FireFlowNet.

Apart from that, some methods exploit the capabilities of certain neuromorphic sensors to produce more than events, such as frames or inertial measurements. In~\cite{Pan2020SingleIO}, the flow of events is employed as a deblurring tool for the frames in highly dynamic scenes, allowing for a better optical flow estimation. In~\cite{Rueckauer2016EvaluationOE}, Rueckauer \textit{et al.} used the IMU integrated in the DAVIS240C camera to determine an exact optical flow estimation for pure rotational movements.

Still, none of these methods has considered the issue of computing optical flow with high-resolution event camera. And, very few of them (\cite{Rueckauer2016EvaluationOE,Akolkar2020SeeBY,ParedesValls2021BackTE}) have been able to achieve real-time compatibility even for low-resolution inputs.

\subsection{Our Orientation: Densifying Events}
Based on the state of the art, methods able to compute high-definition EBOF in real-time are missing. We propose here to treat the event stream through a transformed dense representation\footnote{Notice that we do not need and are not considering using image \emph{reconstruction} methods for computing optical flow. Indeed, while approaches as the ones described in~\cite{Scheerlinck2018ContinuoustimeIE, Stoffregen2020ReducingTS, Rebecq2021HighSA, ParedesValls2021BackTE} do provide interesting reconstruction results, they still show large imperfections, which would degrade the performances of frame-based optical flow methods.} for the following reasons.

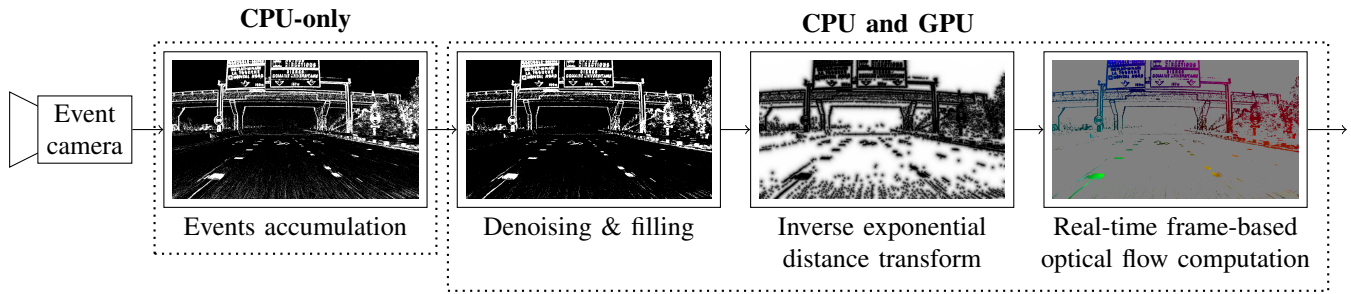
\begin{figure*}
    \centering
    \begin{tikzpicture}[node distance=3.9cm]
        \draw (0,0) -- (-1,0.5) -- (-1,-0.5) -- cycle;
        \node[draw, align=center, fill=white] (EventCamera) {Event\\camera};

        \node[draw, align=center, label={[name=l1] below:Events accumulation}] (Accumulation) [right of=EventCamera, node distance=2.8cm] {\includegraphics[width=0.18\linewidth]{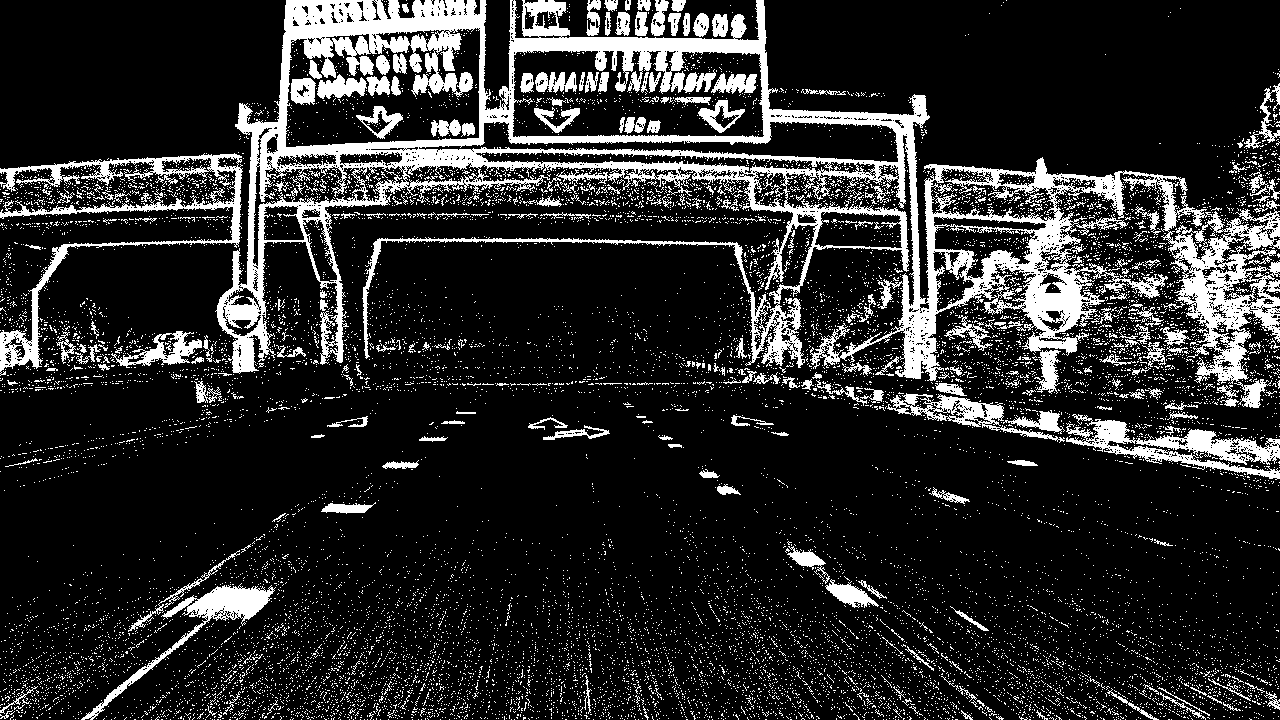}};
        \node[draw, align=center, label={[name=l2] below:Denoising \& filling}] (Denoising) [right of=Accumulation] {\includegraphics[width=0.18\linewidth]{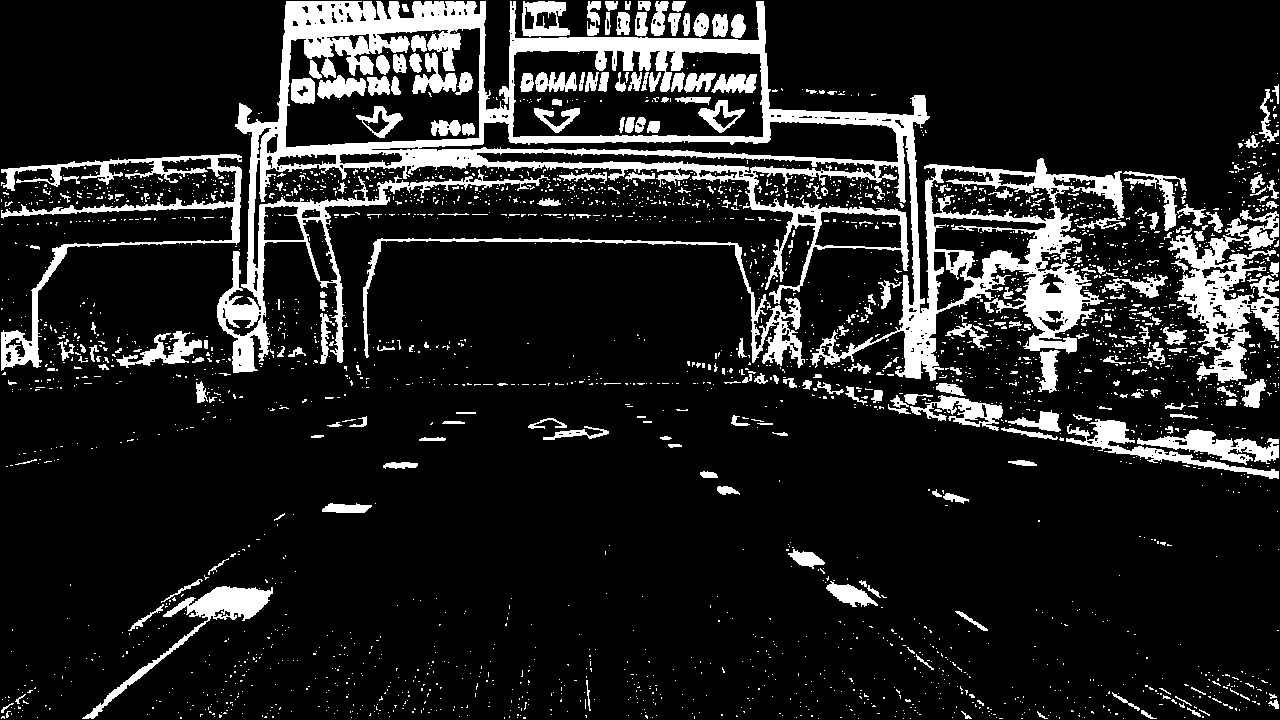}};
        \node[draw, align=center, label={[name=l3, align=center] below:Inverse exponential\\distance transform}] (DistSurf) [right of=Denoising] {\includegraphics[width=0.18\linewidth]{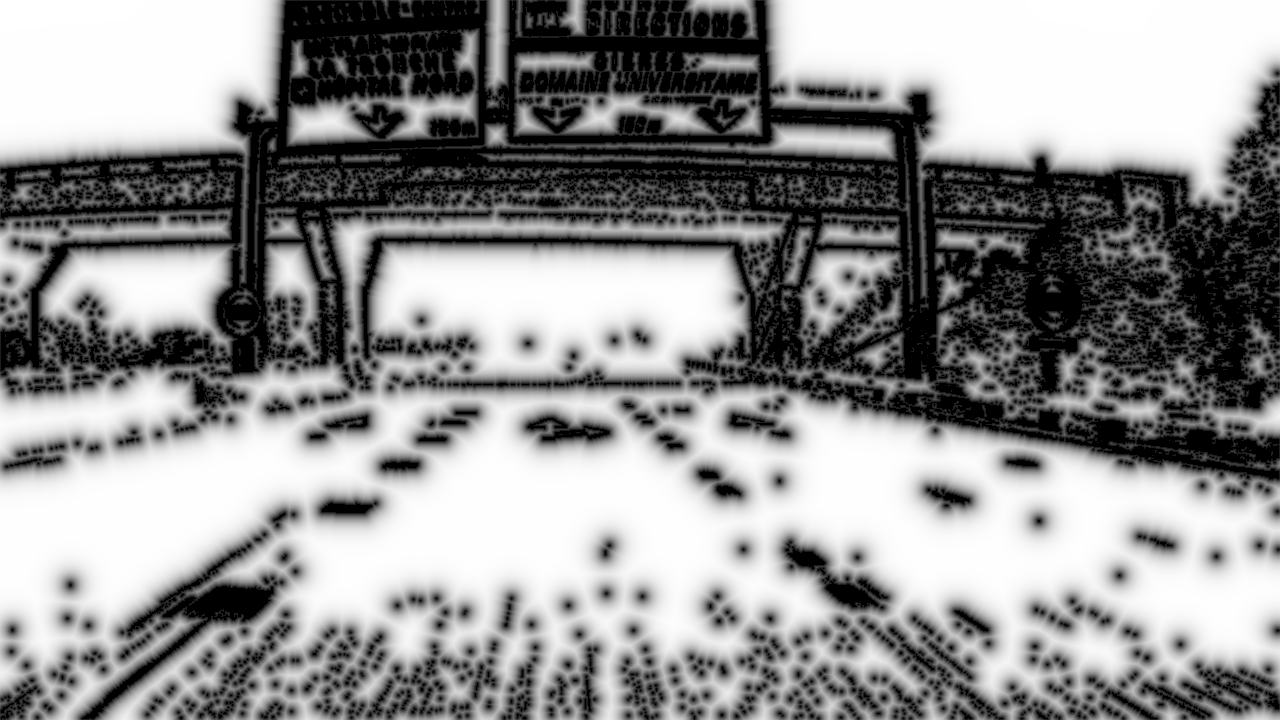}};
        \node[draw, align=center, label={[name=l4, align=center] below:Real-time frame-based\\optical flow computation}] (RTOF) [right of=DistSurf] {\includegraphics[width=0.18\linewidth]{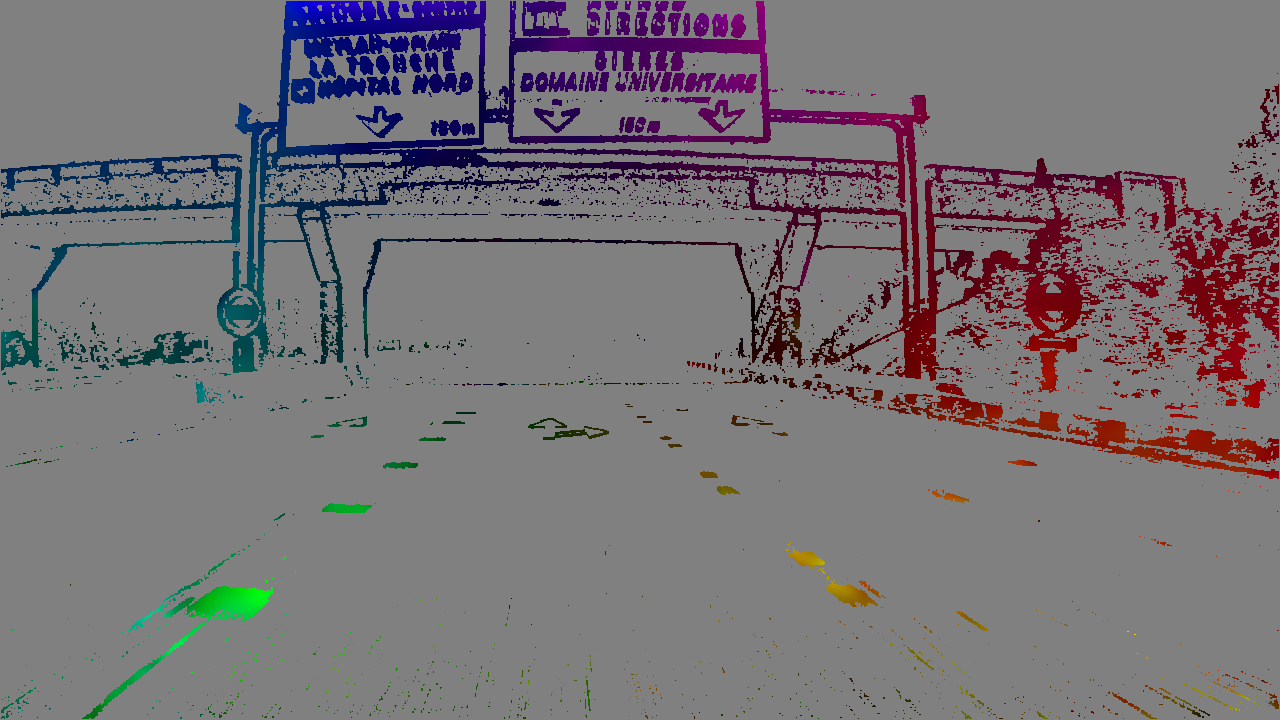}};
        \node[draw=none] (Out) [right of=RTOF, node distance=2.4cm] {};

        \path[->] (EventCamera) edge[above] node[align=center] {} (Accumulation);
        \path[->] (Accumulation) edge[right] node[align=center] {} (Denoising);
        \path[->] (Denoising) edge[above] node[align=center] {} (DistSurf);
        \path[->] (DistSurf) edge[above] node[align=center] {} (RTOF);
        \path[->] (RTOF) edge[above] node[align=center] {} (Out);

        \node[label=above:\textbf{CPU-only},draw,thick,dotted,fit=(Accumulation)(l1)] {};
        \node[label=above:\textbf{CPU and GPU},draw,thick,dotted,fit=(Denoising)(l2) (DistSurf)(l3) (RTOF)(l4)] {};
    \end{tikzpicture}
    \caption{Our event-based optical flow (EBOF) computation architecture, able to run in real-time with low- and high-resolution event cameras. Due to the pipeline architecture, all four blocks are independent parallel processes. Each block depicts the result it produces, for a sample driving sequence.}\label{fig:architecture}
\end{figure*}

While the techniques detailed in the first paragraph of Section~\ref{sec:related_work:eb_flow} do treat each event independently, they result in computationally expensive solutions, unable to reach real-time performances on standard CPUs and GPUs, as they perform an update stage for each new incoming event. This design model becomes even less viable for high-resolution neuromorphic cameras, which produce a much larger event throughput~\cite{Gallego2020EventbasedVA}. Furthermore, the information brought by each new event independently is very little; condensing them over short time windows allows for richer spatial updates~\cite{Gallego2020EventbasedVA}. Building image-like structures from events also facilitates the use of GPUs for fast parallel computations, which compensates the slight latency induced by the short duration of events accumulation. In addition, it allows for the use of state-of-the-art frame-based algorithms, leveraging the decades of research on traditional cameras. Finally, the qualities of event cameras remain. The accumulation time can be accurately adjusted thanks to the high time precision of the events, and the high dynamic range of these sensors allows to operate both for daytime and nighttime~\cite{Gallego2020EventbasedVA}.

At the core of this problem, however, lies the choice of this dense frame-like representation. Accumulating the events during a short time window results only in an image of the scene edges, too simple for adequate optical flow computation. Proposing a densification method for these edge images is therefore a key issue for being able to use frame-based optical flow methods. Moreover, event-based cameras are noisy sensors; a filtering procedure is consequently required for ensuring the stability of these dense frame-like representations.

\section{A Flexible Architecture}\label{sec:architecture}
Following the problem formulation and the reason for the use of a dense representation to reach real-time performances, we detail in this section the novel framework we developed for computing real-time EBOF.

In order to optimize computational time and reach real-time performances, we propose the use of parallelized tasks through a pipeline architecture~\cite{Ramamoorthy1977PipelineA}. An illustration of this framework with example results of each step is available in Fig.~\ref{fig:architecture}. The following subsections will therefore detail how each block contributes towards obtaining the real-time EBOF.

\subsection{Accumulation for Edge Images}
The first component of our architecture is responsible for receiving and accumulating the events from the camera, in short temporal windows, to form ``edge images''. These binary matrices indicate whether or not each pixel produced at least one event during the accumulation time \(\Delta T\). By doing so, each edge image depicts a binary representation of the main edges of the moving objects in the visual scene, which can be used as a first stable medium for computing optical flow.

These edge images do not take into account the polarity of the events: as argued by Almatrafi \textit{et al.}~\cite{Almatrafi2020DistanceSF}, and as we have experimented, both positive and negative events represent similarly the edges of the objects in the visual scene. The additional computational cost linked to treating polarities separately would be too expensive for little improvement in the final results. The choice of \(\Delta T\) is also important and linked to the application: taking a too short \(\Delta T\) will lead to edge images with too few events, resulting in an unstable appearance, while taking a too long \(\Delta T\) will fail to capture clearly the movement of the objects by introducing blur.

Compared to other dense formulations from the literature (time surface~\cite{Delbrck2008FramefreeDD,Nagata2021OpticalFE}, motion-compensated images~\cite{Gallego2018AUC}, reconstructed images~\cite{Rebecq2021HighSA}), our formulation has the benefit of keeping only the information required for frame-based optical flow estimation. Computationally speaking, this makes this solution extremely efficient, as each received event only needs to be placed in a buffer structure. In parallel, a second thread, triggered when the time window has expired, is responsible for collecting all the events from the buffer and creating the edge image, which is then sent for further processing.

\begin{algorithm}[t]
    \caption{Denoising}\label{alg:denoising}
    \SetKwInOut{Input}{Inputs}
    \Input{An edge image \(E\)\\The denoising threshold \(N_d\)}
    \KwOut{The denoised edge image \(E_d\)}
    \(E_d \gets E\)\;
    \ForEach{pixel index \(p \in E\)}{
        \If{\(E[p]\) is an edge pixel}{
            \(n_d \gets\) count of edge pixels among the 4 direct neighbour pixels of \(p\) in \(E\)\;
            \If{\(n_d < N_d\)}{
                \(E_d[p] \gets\) not an edge pixel anymore\;
            } 
        }
    }
\end{algorithm}

\begin{algorithm}[t]
    \caption{Filling}\label{alg:filling}
    \SetKwInOut{Input}{Inputs}
    \Input{A denoised edge image \(E_d\)\\The filling threshold \(N_f\)}
    \KwOut{The denoised and filled edge image \(E_{df}\)}
    \(E_{df} \gets E_d\)\;
    \ForEach{pixel index \(p \in E_d\)}{
        \If{\(E_d\)[p] is not an edge pixel}{
            \(n_f \gets\) count of edge pixels among the 4 direct neighbour pixels of \(p\) in \(E_d\)\;
            \If{\(n_f \geq N_f\)}{
                \(E_{df}[p] \gets\) becomes an edge pixel\;
            }
        }
    }
\end{algorithm}

\begin{figure*}
    \centering
    \begin{tabular}{cccc}
        \includegraphics[width=0.225\linewidth]{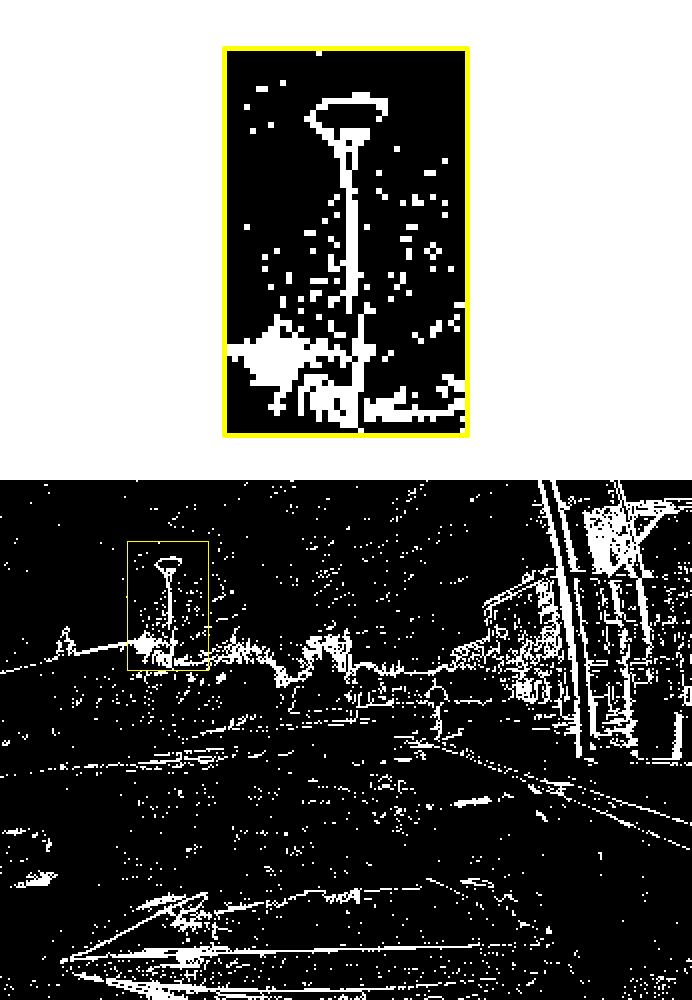} &
        \includegraphics[width=0.225\linewidth]{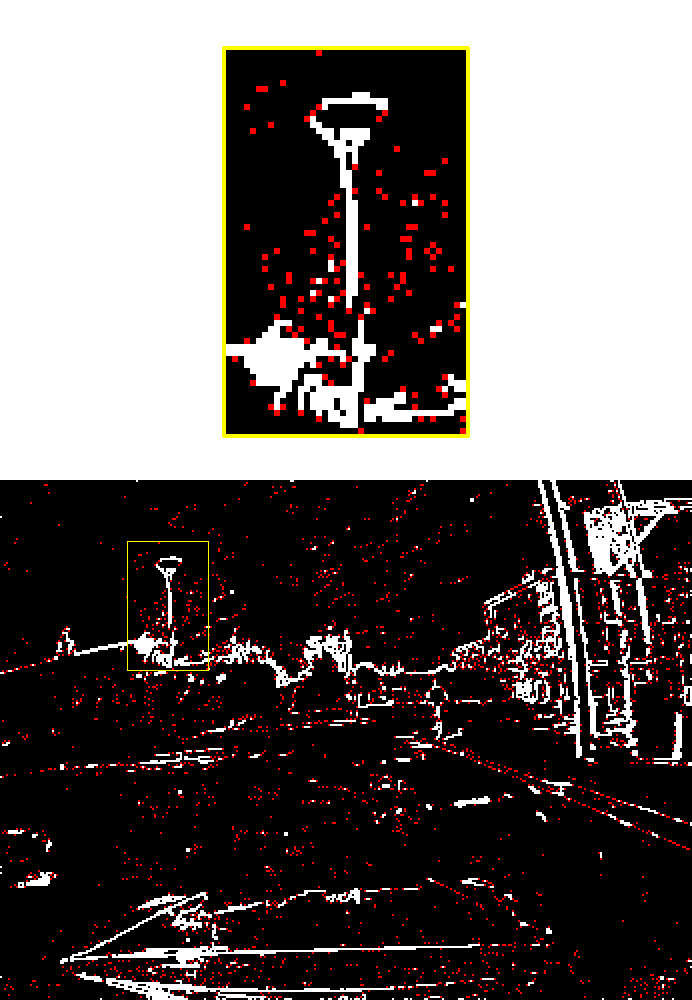} &
        \includegraphics[width=0.225\linewidth]{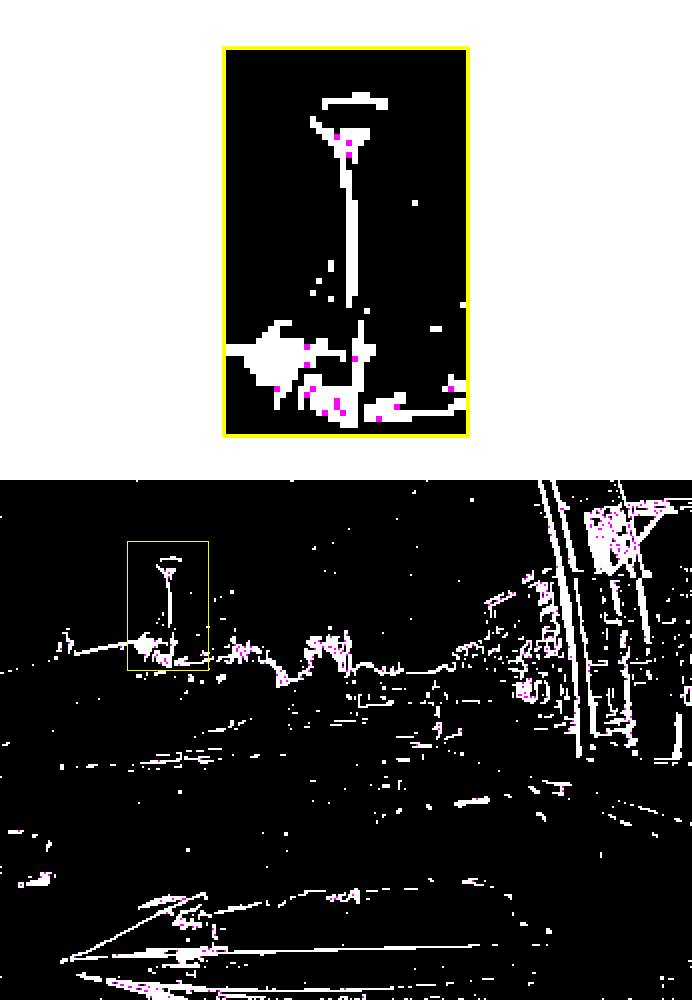} &
        \includegraphics[width=0.225\linewidth]{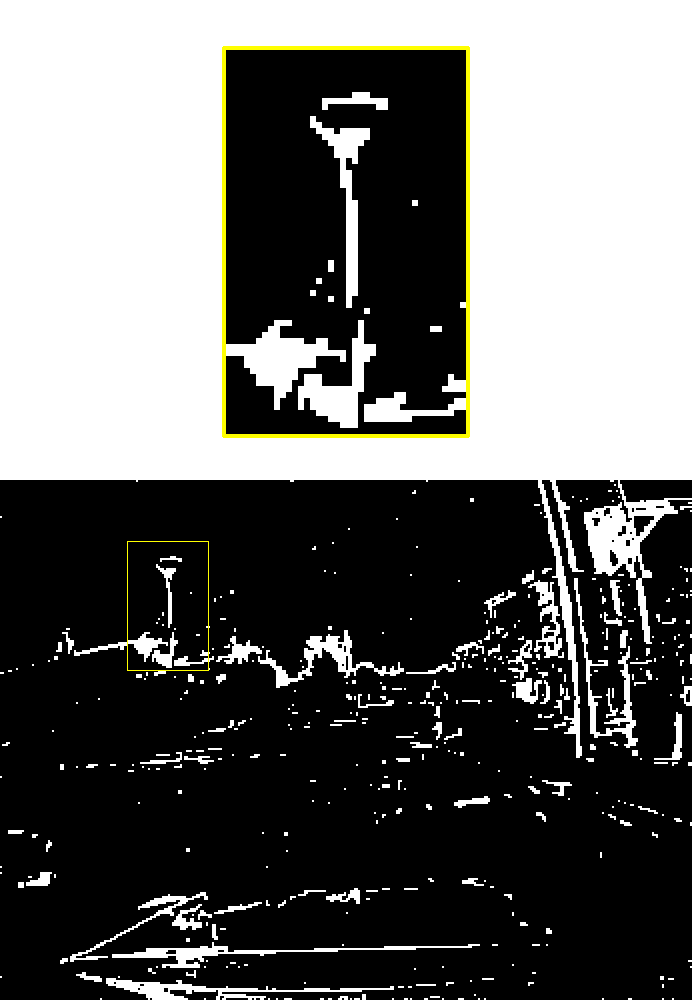}
    \end{tabular}
    \caption{Steps of the denoising and filling process, for a noisy edge image. From left to right: the original noisy edge image; the same image with edge pixels identified as noise in red (\(N_d = 2\)); the denoised edge image with the newly added pixels for the filling in pink (\(N_f = 3\)); the final denoised and filled edge image. A zoomed view of the street lamp (squared in yellow) is also provided for better visibility. Best viewed in color.}\label{fig:denoising_filling}
\end{figure*}

\subsection{Denoising and Filling}\label{ssec:denoisingfilling}
Event cameras generate a significant amount of noise, impacting the quality and stability of the edge images, which in return affects the final optical flow computation.

A solution could be to use one of the state-of-the-art denoising solutions of the literature~\cite{Delbrck2008FramefreeDD,Feng2020EventDB} during the accumulation step, that is, before creating the edge image. Doing so, however, would be computationally expensive, as the sparse and asynchronous nature of the events at that step makes hard to look for neighbour pixels states. Furthermore, many of these solutions were designed for low-resolution sensors, and translate difficultly to higher definition ones.

To circumvent this issue, we propose in this work a novel, fast yet efficient, method for discarding incorrect events. Our approach relies on applying denoising after the edge image creation. Proposed process is then similar to image filtering on the edge map and is computed in two steps: denoising and filling. In the first one, described in Algorithm~\ref{alg:denoising}, erroneous edge pixels are sought to be eliminated, by removing isolated events. The second step, on the contrary, aims at filling locations where an edge pixel is missing, but should have been produced by the camera, in order to help stabilizing the edge images. This process is further described in Algorithm~\ref{alg:filling}. An illustration of both these steps is given in Fig.~\ref{fig:denoising_filling}.

We underline here the importance of computing denoising and filling separately in this order, to avoid creating inconsistencies. Indeed, if the filling step was processed simultaneously with the denoising, then pixels that would later be discarded as noise could contribute to creating incorrect filling pixels, thus introducing new noise.

Denoising and filling thresholds, respectively \(N_d\) and \(N_f\), depend on the event camera configuration, as it may give different noise profile. The aim of the denoising is to discard isolated pixels, that is, pixels with very few neighbours: \(N_d = 1 \text{ or } 2\) appear therefore as the best options. As can be seen in Algorithm~\ref{alg:denoising}, setting \(N_d = 0\) disables the denoising. Then, the goal of the filling is to slightly stabilize the appearance of the edge images, by adding edge pixels in locations where there are enough neighbouring edge pixels to be confident that an edge pixel should have been produced: values of \(N_f = 4 \text{ or } 3\) are therefore the best compromise to add such pixels. As can be seen in Algorithm~\ref{alg:filling}, setting \(N_f = 5\) disables the filling. A general advice is to select \(N_d < N_f\). A sensitivity analysis on \(N_d\) and \(N_f\) is done in Section~\ref{ssec:sensitivity}.

Finally, while this formulation tends to remove small details from the edge images by considering them as noise (as can be seen for instance for the buildings on the right side of the edge images of Fig.~\ref{fig:denoising_filling}), it actually helps obtaining more stable images, by extracting the main edges from the scene, and discarding superfluous textures.

Another advantage of this formulation lies in its simple and parallelizable formulation (the computation for each pixel is independent from the one of its neighbours). We implemented it using the GPU, to exploit its capabilities, and to relieve the CPU, so that it can undertake other complex tasks.

\subsection{The Inverse Exponential Distance Surface}\label{sec:inv_exp_dist_surf}
As the edge images are binary matrices (still after denoising and filling), they can hardly be used as is for computing optical flow with traditional frame-based algorithms. In order to make them viable for frame-based optical flow computation, densifying them through the use of the distance transform, as proposed by Almatrafi \textit{et al.}~\cite{Almatrafi2020DistanceSF}, is an interesting baseline.

\begin{figure}
    \centering
    \begin{tabular}{ccc}
        & & Inverse exp. \\
        Edge image & Distance surface & distance surface \\
        \includegraphics[width=0.284\linewidth]{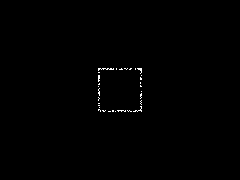} &
        \includegraphics[width=0.284\linewidth]{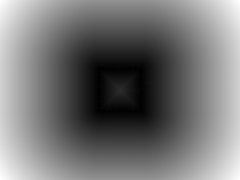} &
        \includegraphics[width=0.284\linewidth]{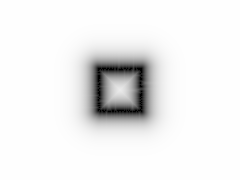} \\
        \includegraphics[width=0.284\linewidth]{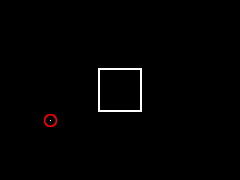} &
        \includegraphics[width=0.284\linewidth]{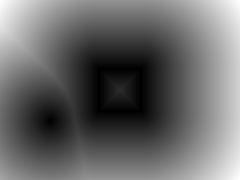} &
        \includegraphics[width=0.284\linewidth]{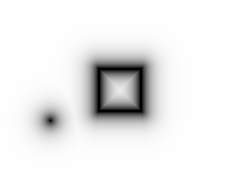} \\
        \includegraphics[width=0.284\linewidth]{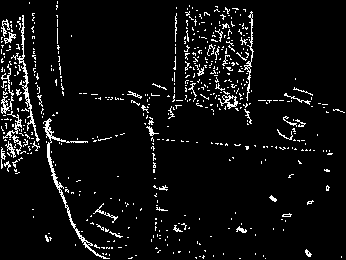} &
        \includegraphics[width=0.284\linewidth]{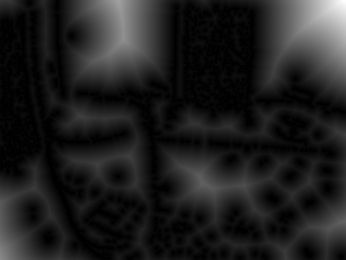} &
        \includegraphics[width=0.284\linewidth]{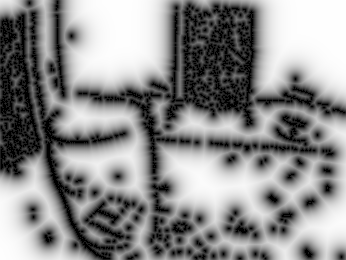} \\
    \end{tabular}
    \caption{Comparison between the original distance surface and proposed inverse exponential distance one (with \(\alpha=2\)). From top to bottom: a simulated square with 50\% of its pixels randomly removed, the same square with an added single pixel of noise (circled in red), and an indoor flying scene from the MVSEC dataset~\cite{Zhu2018TheMS}.}\label{fig:distance_surfaces}
\end{figure}

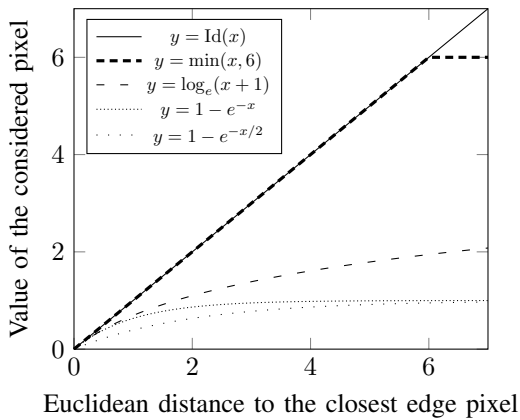
\begin{figure}
    \centering
    \begin{tikzpicture}
        \begin{axis}[width=.8\linewidth, xmin=0, xmax=7, ymin=0, ymax=7, xlabel={Euclidean distance to the closest edge pixel}, ylabel={Value of the considered pixel}, legend style={nodes={scale=0.7, transform shape}}, legend pos=north west, samples=200]
            \addplot[domain=0:7] {x};
            \addplot[domain=0:7, densely dashed, very thick] {min(x, 6)};
            \addplot[domain=0:7, loosely dashed] {ln(x+1)};
            \addplot[domain=0:7, densely dotted] {1-exp(-x)};
            \addplot[domain=0:7, loosely dotted] {1-exp(-x/2)};
            \addlegendentry{\(y = \func{Id}(x)\)}
            \addlegendentry{\(y = \min(x, 6)\)}
            \addlegendentry{\(y = \log_e(x+1)\)}
            \addlegendentry{\(y = 1-e^{-x}\)}
            \addlegendentry{\(y = 1-e^{-x/2}\)}
        \end{axis}
    \end{tikzpicture}
\caption{Values of the distance transform as a function of the distance to the closest edge pixel. Curves represent the original distance transform, the same with an upper bound set to \(6\)px, the natural logarithm version, and our inverse exponential formulation, with \(\alpha \) set to \(1\) and \(2\) respectively.}\label{fig:distance_surface_plot}
\end{figure}

However, this approach has the main drawback of needing a near-perfect denoising, as a single noisy event can disrupt the appearance of the whole distance surface, as shown in the second row of Fig.~\ref{fig:distance_surfaces}. The computed distances are not bounded, meaning that the area of influence of each edge pixel can be infinite, and depends on the presence of other close neighbours. An answer to this problem could be to introduce an upper limit to the computed distances, to restrict the influence of an edge pixel to a fixed neighbourhood. This solution, however, would introduce a non-smooth transition in the distance transform function. It can become an issue for the gradient computation on distance surfaces, often used as part of the optical flow estimation.

Another issue of the approach of Almatrafi \textit{et al.} also appears when distinct objects come close to each other: their edges tend to merge together in the resulting distance surface, making the individual objects indistinguishable, such as in the last row of Fig.~\ref{fig:distance_surfaces}. This phenomenon can lead to incorrect optical flow results, especially when a block-matching or image warping formulation is employed, due to the lack of texture on the produced image. Giving more emphasis to the pixels directly surrounding the edge pixels would help creating distance surfaces with more prominent object edges, limiting this merge issue. A solution could be to employ a function with a logarithmic shape.

To solve jointly both these issues, we propose in this work a novel inverse exponential distance transform formulation:
\begin{equation}\label{eq:inverse_exp_ds}
    d_\text{exp} = 1-e^{-d_\text{Euc}/\alpha},
\end{equation}
where \(d_\text{exp}\) is the inverse exponential distance, \(d_\text{Euc}\) the Euclidean distance to the closest edge pixel (i.e., the original distance transform), both expressed in pixels, and \(\alpha \) a spreading parameter. Fig.~\ref{fig:distance_surface_plot} compares the aforementioned functions over the distance to the closest edge pixel. As can be seen through the plot, the main advantage of our inverse exponential formulation is that, while close to a logarithmic formulation, each edge pixel also has a restricted influence area, after which the values saturate to a value of \(1\).

The spreading parameter \(\alpha \) can be used to determine the size of the neighbourhood influenced by each edge pixel. This parameter conditions the appearance of the distance surface, and regulates the remaining imperfections of the edge images. A low value for \(\alpha \) restricts the area of influence of each edge pixel to only its close neighbours. It limits the influence of the noise on its appearance, but makes the distance surface less stable and more prone to variations. On the contrary, selecting a higher value for \(\alpha \) has the opposite effect: variations of the appearance of the various objects in the scene are well compensated, but noisy events have a more important effect. \(\alpha\) can be rewritten from equation~(\ref{eq:inverse_exp_ds}) as a function of the wanted distance of saturation for \(d_\text{Euc}\), namely, \(d_\text{sat}\), in pixels:
\begin{equation}\label{eq:alpha_fct_d_sat_eps}
    \alpha = -\frac{d_\text{sat}}{\ln(\varepsilon)},
\end{equation}
with \(\varepsilon = 1-d_\text{exp}\), \(\varepsilon > 0\). \(\varepsilon \) is formulated so as to represent the gap between \(d_\text{exp}\) and the saturation value of \(1\). Saturation is therefore reached when this gap \(\varepsilon \) is as small as possible, i.e., \(\varepsilon \to 0\). Since we work in the discrete domain, if the inverse exponential distance surface is coded on 8 bits (values ranging from 0 to 1 are represented by values between 0 and 255), then saturation is reached when \(\varepsilon = \frac{1}{255}\). Integrating this value in (\ref{eq:alpha_fct_d_sat_eps}) results in the final formulation of \(\alpha \) as a function of \(d_\text{sat}\):
\begin{equation}\label{eq:alpha_fct_d_sat}
    \alpha \approx \frac{d_\text{sat}}{5.541}.
\end{equation}
The sensitivity of \(d_\text{sat}\) is studied in the analysis Section~\ref{ssec:sensitivity}.

The interest of our inverse exponential distance surface formulation is illustrated in Fig.~\ref{fig:distance_surfaces}, a side-by-side visual comparison with the original distance surface. Inverse exponential formulation compensates the missing pixels similarly to the original distance surface, while limiting the impact of noisy edge pixels. Also, this formulation displays more distinct edges and keeps objects texture, especially visible in the real complex scene represented in the last row (note how the board and the barrel keep a clear appearance using inverse exponential formulation, while they are hardly distinguishable with the original distance surface).

Regarding the implementation of the distance transform, we employed the fast solution described by Coeurjolly \textit{et al.}~\cite{Coeurjolly2007GomtrieDE}, slightly modified to incorporate our inverse exponential formulation. The choice of this method was especially motivated by its optimized formulation, allowing for large parallelization; this block of our pipeline was therefore implemented on GPU.

\subsection{Selected Frame-Based Optical Flow}
The final block of our architecture is the computation of the optical flow itself. Since the previous steps led to dense image-like structures from the flow of events, any state-of-the-art frame-based optical flow method could be used here.

Within the scope of this article, we selected the approach of Adarve \textit{et al.}~\cite{Adarve2016AFF}. Their method is based on an update-prediction architecture, similar to the one of Black~\cite{Black1992RobustIO}, which predicts optical flow using an image warping process, and temporally propagates the optical flow estimations using an incremental framework. Multiple update-prediction loops are stacked as a pyramidal structure, enabling the capture of both large and fine displacements in the images. Their method was designed for a fast and accurate estimation of the optical flow field, and is implemented on GPU.

The choice of this method as our optical flow computation solution was mainly guided by their use of a predictive filter-based formulation, which, beyond enabling real-time compatibility on GPU, allows for a temporal smoothing of the flow. This property brings stability and robustness to the overall optical flow, given its memory effect, which is beneficial given the sometimes unstable nature of events.

Finally, while this method returns a dense optical flow, covering the whole distance surface image, we then restrict it to the edge pixels of the denoised edge image. Indeed, by nature, events encode sparse information, detailing the pixels for which a change in luminosity was observed. The densification produced by the use of the inverse exponential distance transform differs from an inference of missing data. It is only intended for creating texture and smooth transitions, which are necessary to determine the optical flow.

\section{Evaluation}\label{sec:evaluation}

\subsection{Setup}
The implementation of our method was made using ROS Noetic, in \CC{}11 and CUDA 11.4, combined with the use of the OpenCV 4.2 library. Both the implementation and the evaluation phases were conducted on a HP ZBook 17 G6 laptop, with an Intel i9-9880H CPU, a NVIDIA Quadro RTX 5000 GPU, 64 GB of RAM, and using Ubuntu 20.04.

Regarding the parameters, two configurations were used, respectively for low- and high-resolution input data.

For the low-resolution data (\(346\times260\)) of the MVSEC dataset~\cite{Zhu2018TheMS, Zhu2018EVFlowNetSO}, on one hand, we were restricted to use a temporal window of size \(\Delta T = 1\text{ frame}\)\footnote{In MVSEC, \(\Delta T = 1\text{ frame} \simeq 32\text{ms}\) for ``Indoor flying'' sequences, \(\simeq 22\text{ms}\) for ``Outdoor day'' ones, and \(\simeq 97\text{ms}\) for ``Outdoor night'' ones.} for a fair comparison with the other state-of-the-art methods using this time window~\cite{Zhu2018EVFlowNetSO,Akolkar2020SeeBY,ParedesValls2021SelfSupervisedLO,Lee2020SpikeFlowNetEO,Gehrig2021DenseOF,ParedesValls2021BackTE,Stoffregen2020ReducingTS,Gehrig2019EndtoEndLO,Zhu2019UnsupervisedEL,Ye2020UnsupervisedLO}. For the denoising and filling, we set \(N_d = 1\) and \(N_f = 4\), due to the high noise in these recordings. The inverse exponential distance transform was configured with \(\alpha = 1.08\) (so that \(d_\text{sat} = 6\text{px}\), see (\ref{eq:alpha_fct_d_sat})). Finally, the optical flow computation library~\cite{Adarve2016AFF} was configured with 3 pyramidal layers, with their regularization weights set respectively to \(50.0\), \(250.0\), and \(500.0\), and with \(50\), \(25\), and \(5\) smooth iterations per layer.

For the high-resolution data (\(1280\times720\)), on the other hand, a temporal window \(\Delta T = 15\text{ms}\) was used, to better capture the movements. \(N_d = 2\) and \(N_f = 3\) were empirically chosen, as the best compromise between removing noise and keeping the main edges. \(\alpha=1.08\) (\(d_\text{sat} = 6\text{px}\)) also proved to be the more adequate, allowing to keep the scene details, while compensating potential imperfections. The optical flow library was configured with 3 layers, with regularization weights all set to \(500.0\), and with \(20\) smooth iterations per layer.

\subsection{Datasets}
As part of the evaluation of the proposed methods, four datasets are going to be used in the following subsections. The first one is the low-resolution MVSEC dataset~\cite{Zhu2018TheMS, Zhu2018EVFlowNetSO}, which is currently the only event vision dataset with real data that includes ground truth optical flow. It will serve as the basis for comparison with other state-of-the-art methods. For high-definition data, three complementary datasets are used: the 1 Megapixel Automotive Detection Dataset~\cite{Perot2020LearningTD}, for a deep evaluation on daily driving sequences; a 20-minute-long driving sequence recorded by Prophesee\footnote{\url{https://www.prophesee.ai}}, for visual comparison with the current frame-based state of the art; and a novel high-speed high-definition event-based indoor dataset we recorded as part of this article, to demonstrate the accuracy of our EBOF even under large motions. A summary of these datasets is given in Table~\ref{tab:datasets_comp}.

\begin{table*}
    \centering
    \caption{Comparison between event-based datasets used in this article.}\label{tab:datasets_comp}
    \resizebox{\linewidth}{!}{
        \begin{tabular}{c c c c c c}
        \toprule
        Dataset & Resolution & Scenes & Ground truth optical flow & Frames available & Conditions \\
        \midrule
        \midrule
        MVSEC & \(346\times240\) (low) & Vehicular, drones & Partial & Yes & Day, night \\
        1 Megapixel Automotive Detection & \(1280\times720\) (HD) & Vehicular\textsuperscript{*} & No & No & Day, varying lighting and weather \\
        20-minute-long driving sequence & \(1280\times720\) (HD) & Vehicular\textsuperscript{*} & No & Yes & Day, single long sequence \\
        Our high-speed event dataset & \(1280\times720\) (HD) & Indoor & No & No & Very fast and erratic motions \\
        \bottomrule
        \multicolumn{6}{p{540pt}}{\textsuperscript{*}Both datasets contain diverse driving environments (city, highway, suburbs, countryside, villages), which implies various traffic density and the presence of pedestrians or other road users (cyclists, etc). As such, they are particularly representative of daily scenarios a driver may encounter.} \\
        \end{tabular}
    }
\end{table*}

\subsection{Evaluation Metrics}
In order to evaluate the quality of our optical flow results, three metrics are used as part of this article.

The first two ones, the percentage of outliers and the Average Endpoint Error (AEE), are traditional optical flow metrics, used for instance in the KITTI benchmark~\cite{Geiger2012AreWR}. The percentage of outliers reports the number of pixels for which the error is above 3px and 5\% of the magnitude of the flow vector. The AEE is a raw error measurement on both orientation and magnitude of the flow, computed as following:
\begin{equation}
    AEE = \frac{1}{N}\sum_{i=1}^{N}|v_i-u_i|,
\end{equation}
where \(N\) is the total number of flow vectors, \(u_i\) the \(i\)\textsuperscript{th} estimated flow vector, and \(v_i\) its ground truth equivalent.

However, to this day, no complex high-resolution event-based dataset with a ground truth for optical flow exists. In order to still leverage high-resolution datasets, for instance Prophesee's 1 Megapixel Automotive Detection dataset~\cite{Perot2020LearningTD}, and to provide a quantitative evaluation of our EBOF results, we adopt the Flow Warping Loss (FWL) metric proposed by Stoffregen \textit{et al.}~\cite{Stoffregen2020ReducingTS}. The principle is to compensate and accumulate each raw event (considering its polarity and timestamp) by its computed optical flow, in order to recreate an image of compensated events at a reference time \(t\). If the optical flow is accurate, compensated events superimpose in the same pixel position, producing sharp edges. The FWL then evaluates the sharpness of the produced image, compared to the one where events are not compensated:
\begin{equation}
    \text{FWL} = \frac{\sigma^2(I_\text{comp})}{\sigma^2(I_\text{uncomp})},
\end{equation}
where \(\sigma^2\) is the image variance function, \(I_\text{comp}\) the flow-compensated image of events, and \(I_\text{uncomp}\) the original uncompensated image. By doing so, a final FWL value greater than \(1\) is sought to be obtained, as it indicates that the computed flow is better than the ``zero flow'' (uncompensated) reference.

Finally, in the EBOF illustrations in the following subsections, and in the videos associated to this article, the pixels where no event was received are colored in medium gray.

\subsection{Ablation Studies}
To show the validity of our contributions, we also conducted evaluations with ablations or distance surface alternatives:
\begin{itemize}[align=left, leftmargin=*]
    \item[\textit{Ours\textsubscript{NDF}} --] full proposition without denoising and filling;
    \item[\textit{Ours\textsubscript{DS\_L}} --] linear distance transform, \(y = \func{Id}(x)\);
    \item[\textit{Ours\textsubscript{DS\_LB}} --] upper-bound distance transform (set to 6px, equal to the used \(d_\text{sat}\) value with proposed inverse exponential formulation), \(y = \min(x, 6)\);
    \item[\textit{Ours\textsubscript{DS\_Log}} --] logarithmic distance transform, \(y = \log_e(x+1)\).
\end{itemize}
Fig.~\ref{fig:distance_surface_plot} illustrates the shape of these variants.

\subsection{Evaluation on the MVSEC Dataset}\label{sec:eval_mvsec}
We evaluated our EBOF method on the low-resolution (\(346\times260\)) MVSEC dataset proposed by Zhu \textit{et al.}~\cite{Zhu2018TheMS, Zhu2018EVFlowNetSO}. Despite several shortcomings highlighted by its authors --- namely, errors created by moving objects, an approximate synchronization, and the use of default biases  --- this dataset remains the main reference for evaluating EBOF results on complex real-life sequences. Therefore, we present in Table~\ref{tab:mvsec_results} our error measurements on this dataset, compared to other reference methods from the literature (both non real-time and real-time capable). We also compare them to a ``zero flow'' reference, i.e., error measurements when the estimated optical flow is set to a null vector field. Note that, similarly to other authors such as \cite{Zhu2018EVFlowNetSO, Stoffregen2020ReducingTS}, for ``Outdoor'' sequences, we ignored the pixels where the hood of the car is visible. In the dataset, these pixels contain incorrect ground truth values.

\begin{table*}
    \centering
    \caption{Results on the MVSEC dataset. Bold indicates the best results for non real-time and real-time versions separately.}\label{tab:mvsec_results}
    \resizebox{\linewidth}{!}{
        \begin{tabular}{c cc cc cc cc cc cc cc cc}
        \toprule
        Sequence & \multicolumn{2}{c}{Indoor flying 1} & \multicolumn{2}{c}{Indoor flying 2} & \multicolumn{2}{c}{Indoor flying 3} & \multicolumn{2}{c}{Outdoor day 1} & \multicolumn{2}{c}{Outdoor day 2} & \multicolumn{2}{c}{Outdoor night 1} & \multicolumn{2}{c}{Outdoor night 2} & \multicolumn{2}{c}{Outdoor night 3} \\
        & AEE & \% outliers & AEE & \% outliers & AEE & \% outliers & AEE & \% outliers & AEE & \% outliers & AEE & \% outliers & AEE & \% outliers & AEE & \% outliers \\
        \midrule
        \midrule
        Zero flow & 1.71 & 8.9 & 3.03 & 40.2 & 2.53 & 29.1 & 1.46 & 5.1 & 1.70 & 13.0 & 5.41 & 63.8 & 6.62 & 73.7 & 7.2 & 77.1 \\
        \midrule
        \midrule
        \multicolumn{17}{c}{\textbf{Non Real-Time}} \\
        EV-FlowNet\textsubscript{MVSEC}~\cite{Zhu2018EVFlowNetSO} & 1.03 & 2.2 & 1.72 & 15.1 & 1.53 & 11.9 & \textit{0.49}\textsuperscript{*} & \textit{0.2}\textsuperscript{*} & - & - & - & - & - & - & - & - \\
        EV-FlowNet\textsubscript{MVSEC} (updated)\textsuperscript{\textdagger} & 0.85 & 0.9 & 1.29 & 7.5 & 1.13 & 5.3 & 0.56 & 0.4 & - & - & \textbf{1.90} & \textbf{18.6} & \textbf{2.26} & \textbf{21.9} & \textbf{2.13} & \textbf{20.4} \\
        EV-FlowNet\textsubscript{EST}~\cite{Gehrig2019EndtoEndLO} & 0.97 & 0.9 & 1.38 & 8.2 & 1.43 & 6.5 & - & - & - & - & - & - & - & - & - & - \\
        EV-FlowNet\textsubscript{HQF}~\cite{Stoffregen2020ReducingTS} & \textbf{0.56} & 1.0 & \textbf{0.66} & \textbf{1.0} & \textbf{0.59} & \textbf{1.0} & 0.68 & 1.0 & \textbf{0.82} & \textbf{1.0} & - & - & - & - & - & - \\
        EV-FlowNet\textsubscript{DR}~\cite{ParedesValls2021BackTE} & 0.79 & 1.2 & 1.40 & 10.9 & 1.18 & 7.4 & 0.92 & 5.4 & - & - & - & - & - & - & - & - \\
        Zhu \textit{et al.}~\cite{Zhu2019UnsupervisedEL} & 0.58 & \textbf{0.0} & 1.02 & 4.0 & 0.87 & 3.0 & 0.32 & \textbf{0.0} & - & - & - & - & - & - & - & - \\
        Spike-FlowNet~\cite{Lee2020SpikeFlowNetEO} & 0.84 & - & 1.28 & - & 0.87 & - & 0.49 & - & - & - & 2.77\textsuperscript{\textdaggerdbl} & - & - & - & - & - \\
        \textit{Nagata et al.~\cite{Nagata2021OpticalFE}}\textsuperscript{\textsection} & \textit{0.28} & - & \textit{0.42} & - & \textit{0.38} & - & \textit{0.26} & - & \textit{0.35} & - & \textit{0.33} & - & \textit{0.36} & - & \textit{0.36} & - \\
        Spiking EV-FlowNet~\cite{ParedesValls2021SelfSupervisedLO} & 0.60 & 0.5 & 1.17 & 8.1 & 0.93 & 5.6 & 0.47 & 0.2 & - & - & - & - & - & - & - & - \\
        Gehrig \textit{et al.}~\cite{Gehrig2021DenseOF} & - & - & - & - & - & - & \textbf{0.24} & \textbf{0.0} & - & - & - & - & - & - & - & - \\
        \midrule
        \midrule
        \multicolumn{17}{c}{\textbf{Real-Time}} \\
        \textit{ECN\textsubscript{masked}~\cite{Ye2020UnsupervisedLO}}\textsuperscript{\textparagraph} & - & - & - & - & - & - & \textit{0.30} & \textit{0.0} & - & - & \textit{0.53} & \textit{1.1} & \textit{0.49} & \textit{1.0} & \textit{0.49} & \textit{1.1} \\
        Akolkar \textit{et al.}~\cite{Akolkar2020SeeBY} & 1.52 & - & 1.59 & - & 1.89 & - & 2.75 & - & - & - & 4.47 & - & - & - & - & - \\
        FireFlowNet~\cite{ParedesValls2021BackTE} & 0.97 & 2.6 & 1.67 & 15.3 & 1.43 & 11.0 & 1.06 & 6.6 & - & - & - & - & - & - & - & - \\
        Ours & 0.52 & \textbf{0.1} & 0.98 & 5.5 & 0.71 & 2.1 & \textbf{0.53} & \textbf{0.2} & \textbf{0.74} & \textbf{1.2} & \textbf{2.91} & \textbf{30.6} & \textbf{3.45} & \textbf{39.1} & \textbf{3.62} & \textbf{39.8} \\
        Ours\textsubscript{NDF} & \textbf{0.49} & \textbf{0.1} & \textbf{0.92} & \textbf{4.6} & \textbf{0.68} & \textbf{1.6} & 0.54 & 0.4 & 0.75 & 1.3 & 2.99 & 31.8 & 3.56 & 40.4 & 3.70 & 40.9 \\
        Ours\textsubscript{DS\_L} & 1.81 & 16.4 & 2.54 & 26.4 & 1.95 & 18.2 & 2.12 & 21.7 & 1.30 & 8.7 & 4.04 & 45.8 & 4.78 & 55.6 & 5.10 & 58.7 \\
        Ours\textsubscript{DS\_LB} & 0.62 & 0.3 & 1.02 & 5.6 & 0.79 & 1.8 & 0.64 & 0.5 & 0.79 & 1.3 & 3.05 & 32.5 & 3.68 & 41.8 & 3.90 & 43.4 \\
        Ours\textsubscript{DS\_Log} & 0.70 & 1.4 & 1.07 & 6.5 & 0.82 & 2.4 & 0.69 & 1.6 & 0.79 & 1.9 & 3.08 & 33.0 & 3.70 & 42.1 & 3.93 & 43.8 \\
        \bottomrule
        \multicolumn{17}{l}{\textsuperscript{*}The authors of EV-FlowNet only evaluated this sequence on a carefully selected 18-second-long extract.} \\
        \multicolumn{17}{l}{\textsuperscript{\textdagger}The authors of EV-FlowNet provide an updated version of their model (\url{https://github.com/daniilidis-group/EV-FlowNet/}), which we recomputed their results with.} \\
        \multicolumn{17}{l}{\textsuperscript{\textdaggerdbl}This result was computed by ourselves, using the code provided by the authors (\url{https://github.com/chan8972/Spike-FlowNet}).} \\
        \multicolumn{17}{p{695pt}}{\textsuperscript{\textsection}The authors used an accumulation window of \(\Delta t = 2.5\text{ms}\), whereas all the other results presented here used an accumulation window of \(\Delta t = 1\text{ frame}\) (10 to 40 times longer), thus creating an unfair comparison. Therefore, we report their measurements here, but do not consider them for comparison purposes.}\\
        \multicolumn{17}{p{695pt}}{\textsuperscript{\textparagraph}The results of ECN\textsubscript{masked} were obtained after manually removing from the dataset erroneous ground truth values created by independently moving objects. While allowing for better AEE and outliers results, it creates an unfair comparison baseline. Thus, we report their measurements here, but do not consider them for comparison purposes.}
        \end{tabular}
    }
\end{table*}

From these results, we obtain AEEs in the order of one pixel, except for nighttime sequences, where the longer accumulation time of \(\Delta T \simeq 97\text{ ms}\) results in greater magnitudes of errors. Our AEE results are remarkably always close to or even better than all the non-real-time state-of-the-art approaches (EV-FlowNet\textsubscript{HQF}~\cite{Stoffregen2020ReducingTS} notably). We display vastly better results than FireFlowNet~\cite{ParedesValls2021BackTE}, which is our main comparison point when it comes to fast EBOF methods.

Outlier percentages are also very low, only increasing for the nighttime driving scenes. However, as noted by Ye \textit{et al.}~\cite{Ye2020UnsupervisedLO}, MVSEC ground truth flow is valid only for static world; the moving objects, numerous in the nighttime scenes, could not be kept in the reference, creating errors in the ground truth.

When compared to the ablated versions of our method, it can be seen that the ``No denoising'' \textit{Ours\textsubscript{NDF}} performs better for the indoor sequences, where the lighting of the scene is controlled, and the noise therefore less prominent. In that case, the denoising and filling step will mostly tend to eliminate small texture details from the scene, which could in reality be kept to improve the stability of its appearance. On the outdoor sequences, on the contrary, our denoising shows its importance, as the noise generated by the environment becomes much more essential to discard to obtain accurate optical flow results. Regarding the distance surface alternatives, they all display worse results than proposed inverse exponential, both for indoor and outdoor sequences; the original linear distance surface \textit{Ours\textsubscript{DS\_L}}, notably, displays here the worst results, even worse than the zero flow baseline in some sequences.

\begin{table*}
    \centering
    \caption{FWL results on the MVSEC dataset}\label{tab:fwl_mvsec_results}
    \resizebox{\linewidth}{!}{
        \begin{tabular}{c c c c c c c c c}
        \toprule
        Sequence & Indoor flying 1 (cut)\textsuperscript{*} & Indoor flying 2 (cut)\textsuperscript{*} & Indoor flying 3 (cut)\textsuperscript{*} & Outdoor day 1 (cut)\textsuperscript{*} & Outdoor day 2 (cut)\textsuperscript{*} & Outdoor night 1 & Outdoor night 2 & Outdoor night 3 \\
        \midrule
        \midrule
        EV-FlowNet\textsubscript{MVSEC}~\cite{Zhu2018EVFlowNetSO}\textsuperscript{\textdagger} & 1.02 & 1.13 & 1.06 & 1.15 & 1.21 & - & - & - \\
        EV-FlowNet\textsubscript{HQF}~\cite{Stoffregen2020ReducingTS} & 1.14 & 1.36 & 1.23 & \textbf{1.27} & \textbf{1.20} & - & - & - \\
        Ours & \textbf{1.21} & \textbf{1.41} & \textbf{1.37} & 1.17 & 1.17 & \textbf{1.35} & \textbf{1.45} & \textbf{1.42} \\
        Ours\textsubscript{NDF} & 1.16 & 1.34 & 1.29 & 1.15 & 1.14 & 1.28 & 1.37 & 1.34 \\
        Ours\textsubscript{DS\_L} & 1.16 & 1.29 & 1.27 & 1.13 & 1.12 & 1.21 & 1.28 & 1.25 \\
        Ours\textsubscript{DS\_LB} & 1.20 & \textbf{1.41} & 1.35 & 1.16 & 1.17 & 1.34 & 1.43 & 1.39 \\
        Ours\textsubscript{DS\_Log} & \textbf{1.21} & 1.39 & 1.36 & 1.16 & 1.16 & 1.32 & 1.41 & 1.38 \\
        \bottomrule
        \multicolumn{9}{l}{\textsuperscript{*}Stoffregen \textit{et al.}~\cite{Stoffregen2020ReducingTS} selected cuts from these sequences to evaluate their method; we use the same extracts here.} \\
        \multicolumn{9}{l}{\textsuperscript{\textdagger}As computed by Stoffregen \textit{et al.}~\cite{Stoffregen2020ReducingTS}.}
        \end{tabular}
    }
\end{table*}

Despite the presence of a ground truth in MVSEC dataset, we also computed the FWL metric, in Table~\ref{tab:fwl_mvsec_results}. Our results consistently surpass the value of \(1\), indicating an optical flow estimation better than the zero flow reference. Most importantly, they surpass those of EV-FlowNet\textsubscript{MVSEC}~\cite{Zhu2018EVFlowNetSO} and EV-FlowNet\textsubscript{HQF}~\cite{Stoffregen2020ReducingTS} in most of the sequences. While these results may sometimes slightly contrast with those presented in Table~\ref{tab:mvsec_results}, from our understanding, they further underline the inconsistencies in the ground truth flow of the MVSEC dataset, but still demonstrate the high accuracy of our approach compared to non-real-time ones.

Regarding the FWL comparison with the alternative methods, proposed denoised inverse exponential formulation displays the best results for all sequences. The \textit{Ours\textsubscript{NDF}} version consistently performs worse, as the small details in the scene, not discarded as noise here, compensate more difficultly than the main edges and lower the results yielded by the FWL metric. Once again, the alternative distance surface formulations all provide worse results than the inverse exponential one. However, it should be underlined here that the ``linear-bound'' \textit{Ours\textsubscript{DS\_LB}} method displays results closer to the inverse exponential distance surface than anticipated.

\begin{figure*}
    \centering
    \begin{tabular}{cccc}
        Grayscale Image & Edge Image & Ground Truth Flow & Our Flow \\
        \includegraphics[width=0.22\linewidth]{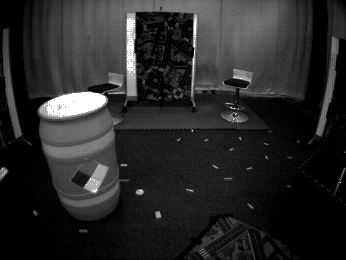} &
        \includegraphics[width=0.22\linewidth]{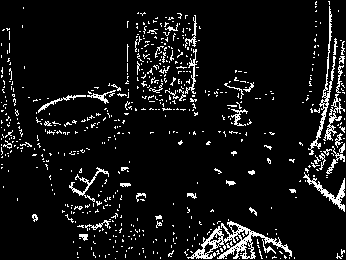} &
        \includegraphics[width=0.22\linewidth]{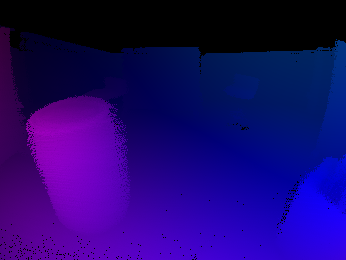} &
        \includegraphics[width=0.22\linewidth]{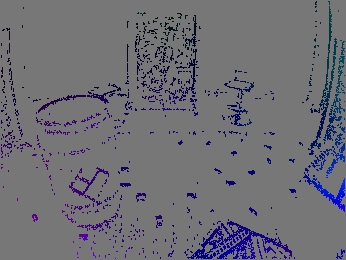} \\ % 22.0s
        \includegraphics[width=0.22\linewidth]{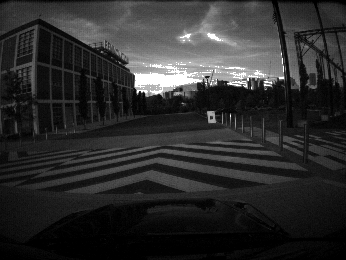} &
        \includegraphics[width=0.22\linewidth]{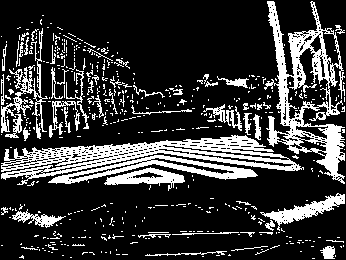} &
        \includegraphics[width=0.22\linewidth]{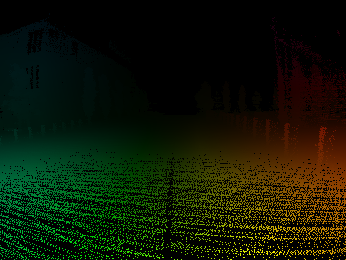} &
        \includegraphics[width=0.22\linewidth]{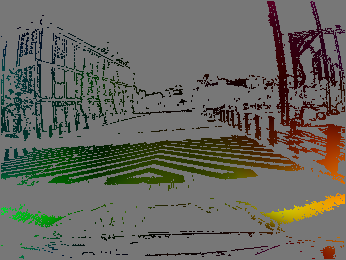} \\ % 88.4s
        \includegraphics[width=0.22\linewidth]{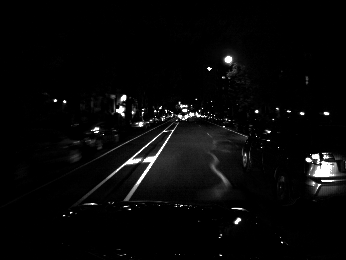} &
        \includegraphics[width=0.22\linewidth]{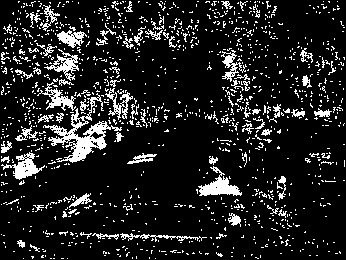} &
        \includegraphics[width=0.22\linewidth]{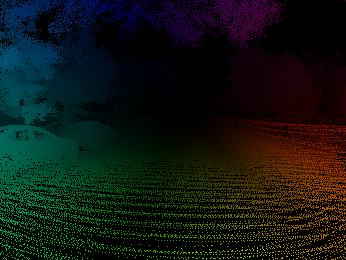} &
        \includegraphics[width=0.22\linewidth]{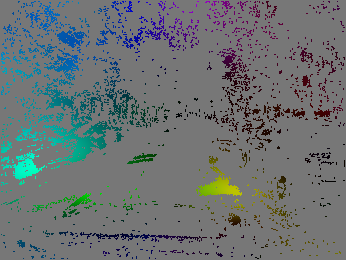} \\ % 235.0s
    \end{tabular}
    \caption{Qualitative results on MVSEC dataset. Sequences, from top to bottom: Indoor flying 1, Outdoor day 1, and Outdoor Night 1. Best viewed in color.}\label{fig:results_mvsec}
\end{figure*}

Finally, we present in Fig.~\ref{fig:results_mvsec} qualitative optical flow results for sequences from the MVSEC dataset. It can be seen that our EBOF is visually close to the reference. Limitations in the ground truth of the dataset can also be observed: the hood of the car is not taken into account in the ground truth of the outdoor sequences (second and third rows), and the moving vehicle at the right of the car in the last row is associated to an incorrectly smoothed ground truth flow.

\begin{table*}
    \centering
    \caption{FWL results on Prophesee's 1 Megapixel Automotive Detection dataset}\label{tab:fwl_1mp_results}
    \resizebox{\linewidth}{!}{
        \begin{tabular}{c c c c c c c c c c c c c c c c}
        \toprule
        Sequence & Jan.\ 30 & Feb.\ 15 & Feb.\ 18 & Feb.\ 19 & Feb.\ 21 & Feb.\ 22 & Apr.\ 12 & Apr.\ 18 & Jun.\ 11 & Jun.\ 14 & Jun.\ 17 & Jun.\ 19 & Jun.\ 21 & Jun.\ 26 & \textit{Global} \\
        \midrule
        \midrule
        Ours & \textbf{1.63} & 1.47 & \textbf{1.34} & \textbf{1.64} & \textbf{1.36} & \textbf{1.51} & 1.14 & 1.54 & \textbf{1.80} & \textbf{1.35} & \textbf{1.46} & \textbf{1.38} & \textbf{1.54} & 1.56 & \textbf{1.46} \\
        Ours\textsubscript{NDF} & 1.43 & 1.37 & 1.25 & 1.52 & 1.28 & 1.41 & 1.08 & 1.43 & 1.63 & 1.27 & 1.35 & 1.31 & 1.46 & 1.46 & 1.37 \\
        Ours\textsubscript{DS\_L} & 1.44 & 1.40 & 1.32 & 1.41 & 1.29 & 1.38 & \textbf{1.21} & 1.27 & 1.50 & 1.28 & 1.27 & 1.30 & 1.32 & 1.42 & 1.34 \\
        Ours\textsubscript{DS\_LB} & \textbf{1.63} & \textbf{1.48} & 1.33 & 1.63 & 1.35 & 1.49 & 1.14 & \textbf{1.55} & 1.78 & 1.34 & 1.43 & 1.37 & 1.52 & \textbf{1.57} & 1.45 \\
        Ours\textsubscript{DS\_Log} & 1.60 & 1.47 & 1.33 & 1.60 & 1.34 & 1.47 & 1.14 & 1.54 & 1.75 & 1.33 & 1.41 & \textbf{1.38} & 1.51 & 1.56 & 1.43 \\
        \bottomrule
        \end{tabular}
    }
\end{table*}

\subsection{Evaluation on the 1 Mp Automotive Detection Dataset}\label{sec:eval_prophesee_1mp}
The arrival of high-resolution neuromorphic cameras means that a more thorough evaluation including these new sensors has to be conducted. In the context of this article, the 1 Megapixel Automotive Detection dataset~\cite{Perot2020LearningTD} from Prophesee --- while initially intended for automotive object recognition purposes --- appears as the most complete, publicly available high-definition baseline for conducting our evaluation. Given its density (1.2 TB, 14 hours of data), we settled on the use of only its ``test'' sequences for the evaluation, which account for a total of 2 hours of raw data.

We tried to adapt the codes of the low-resolution EV-FlowNet methods proposed by Zhu \textit{et al.}~\cite{Zhu2018EVFlowNetSO} and Stoffregen \textit{et al.}~\cite{Stoffregen2020ReducingTS}, to provide the same comparisons as on the MVSEC dataset. However, the results it yielded are not representative of the qualities of the methods, as their neural-network-based approaches were not designed nor trained for high-resolution input data. The most apparent and limiting issues are the computation times (the code was not optimized for high-resolution data), and the absence of denoising (it created large artifacts in the optical flow results).

\begin{figure*}
    \centering
    \begin{tabular}{cccc}
        Reference Image & Edge Image & Image-based Flow (RAFT) & Our Flow \\
        \includegraphics[width=0.22\linewidth]{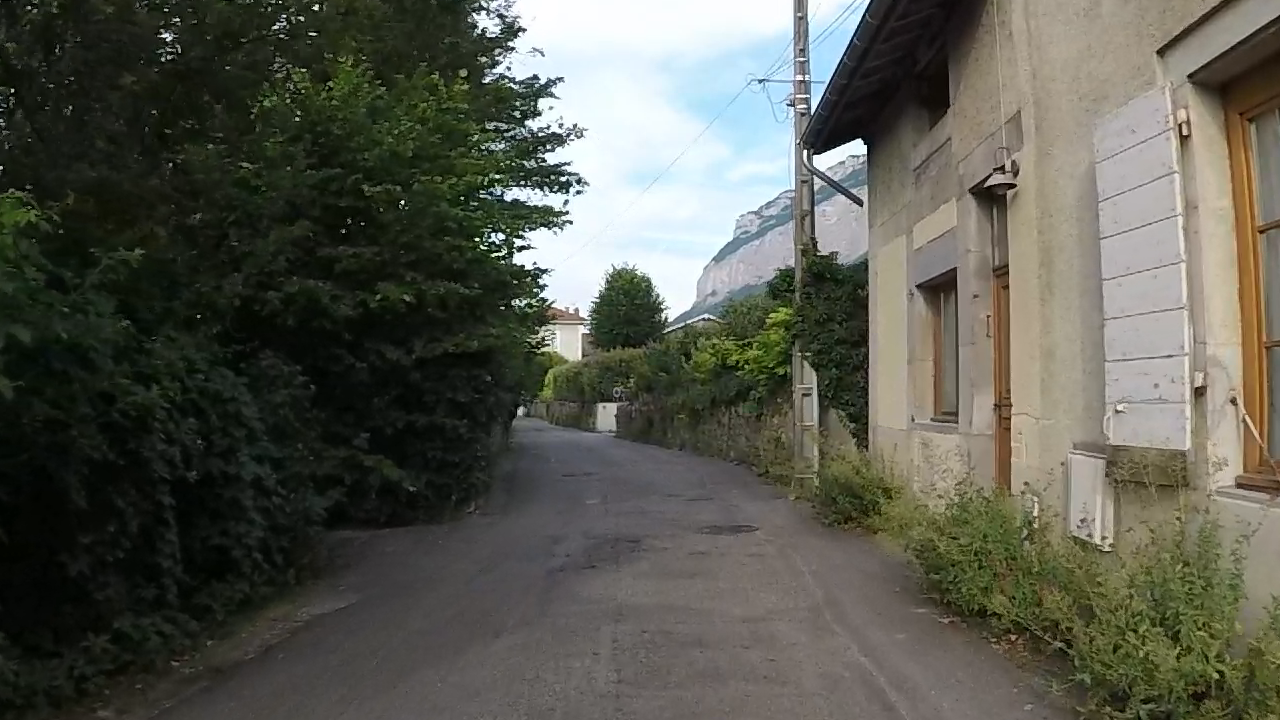} &
        \includegraphics[width=0.22\linewidth]{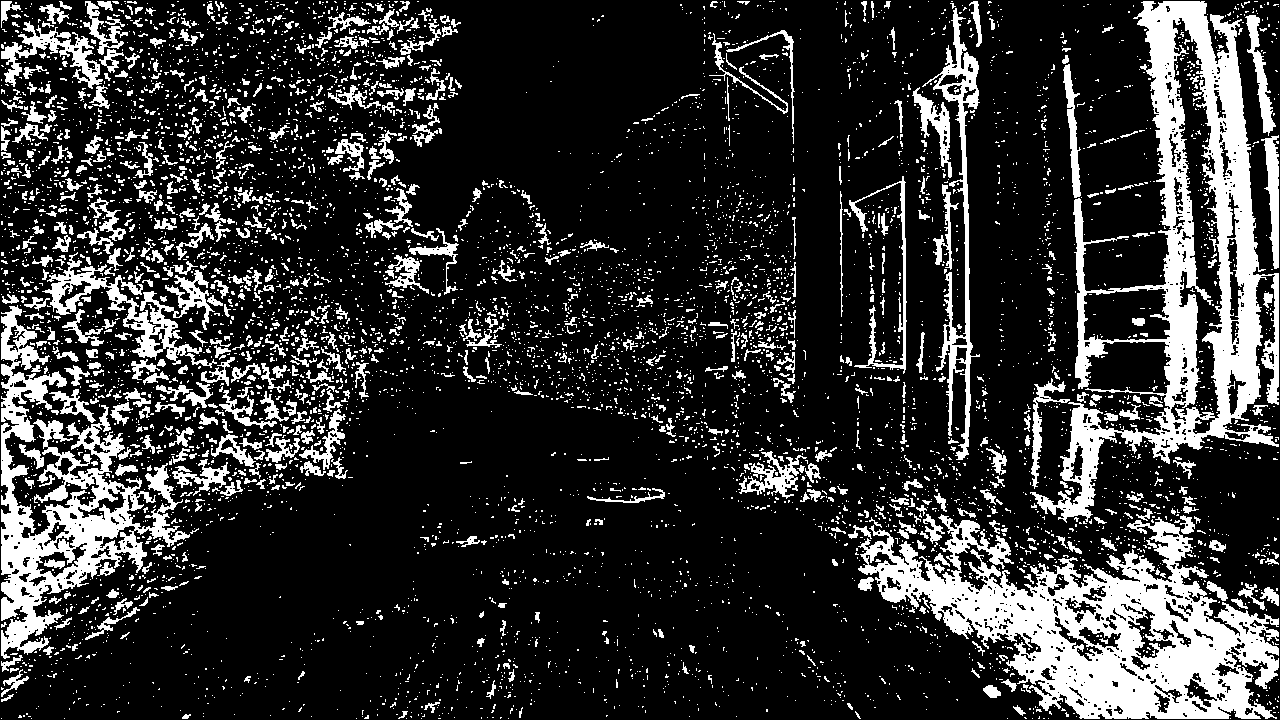} &
        \includegraphics[width=0.22\linewidth]{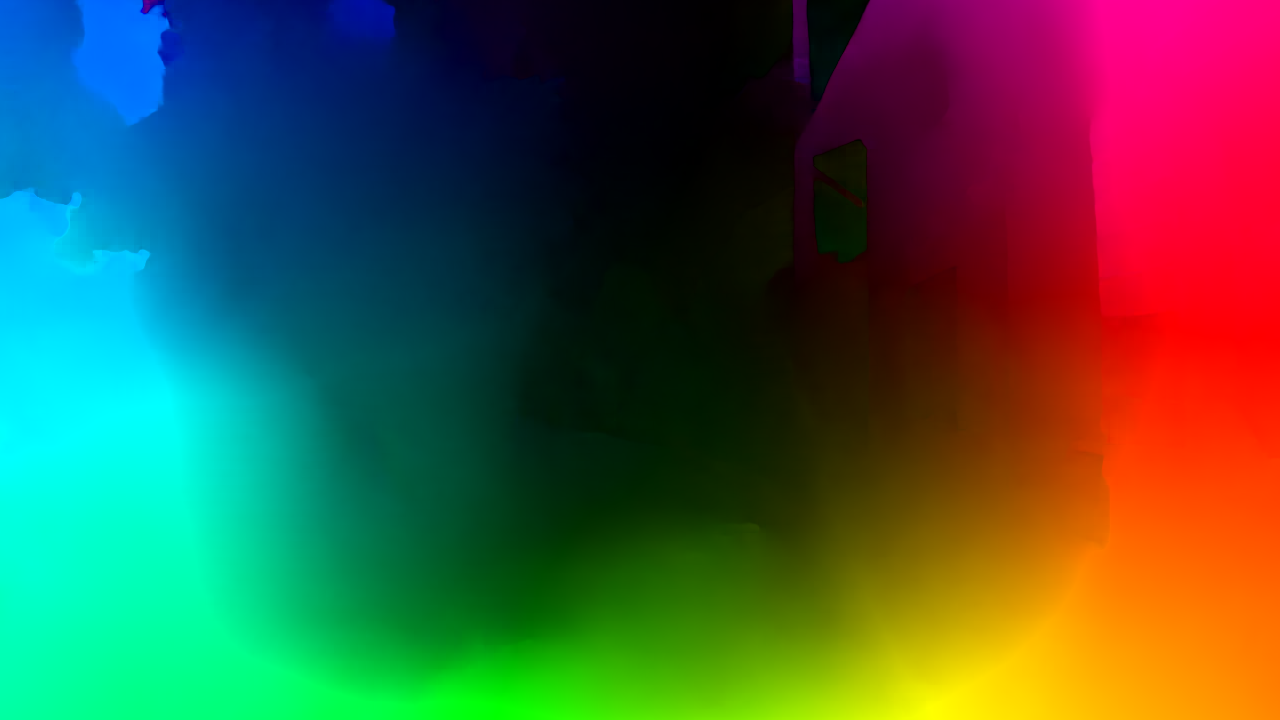} &
        \includegraphics[width=0.22\linewidth]{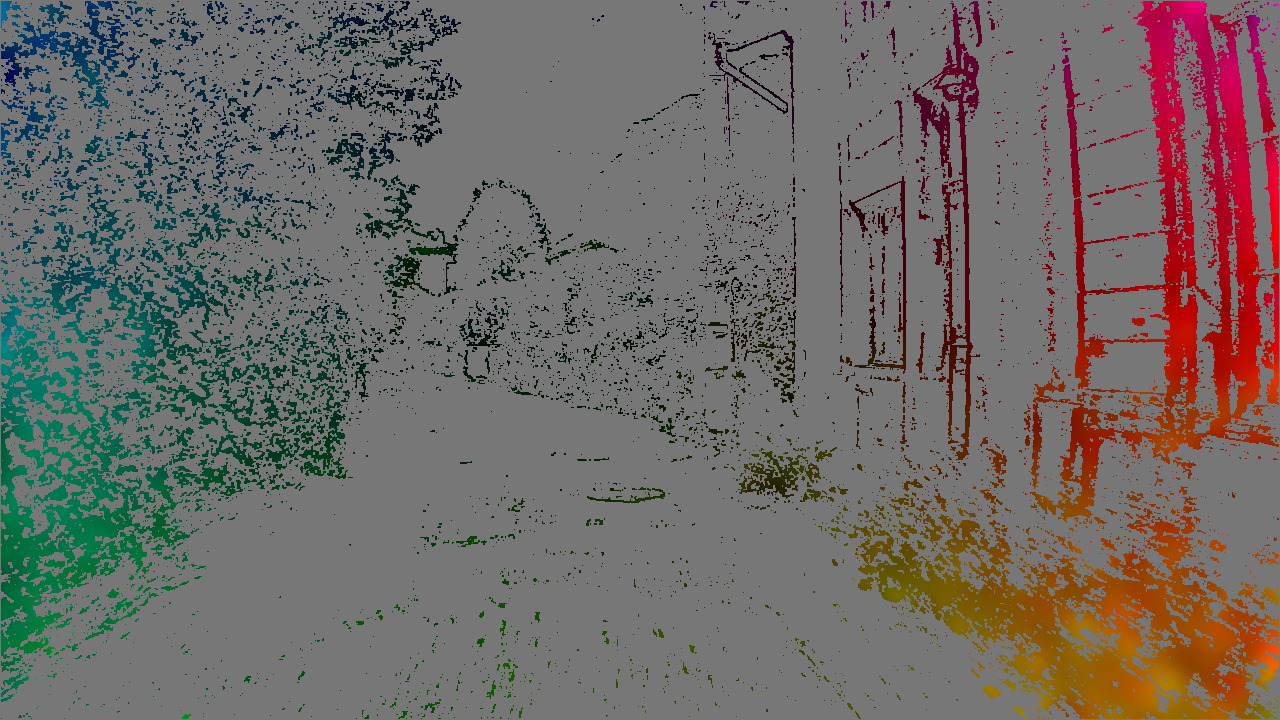} \\ % 27.0s
        \includegraphics[width=0.22\linewidth]{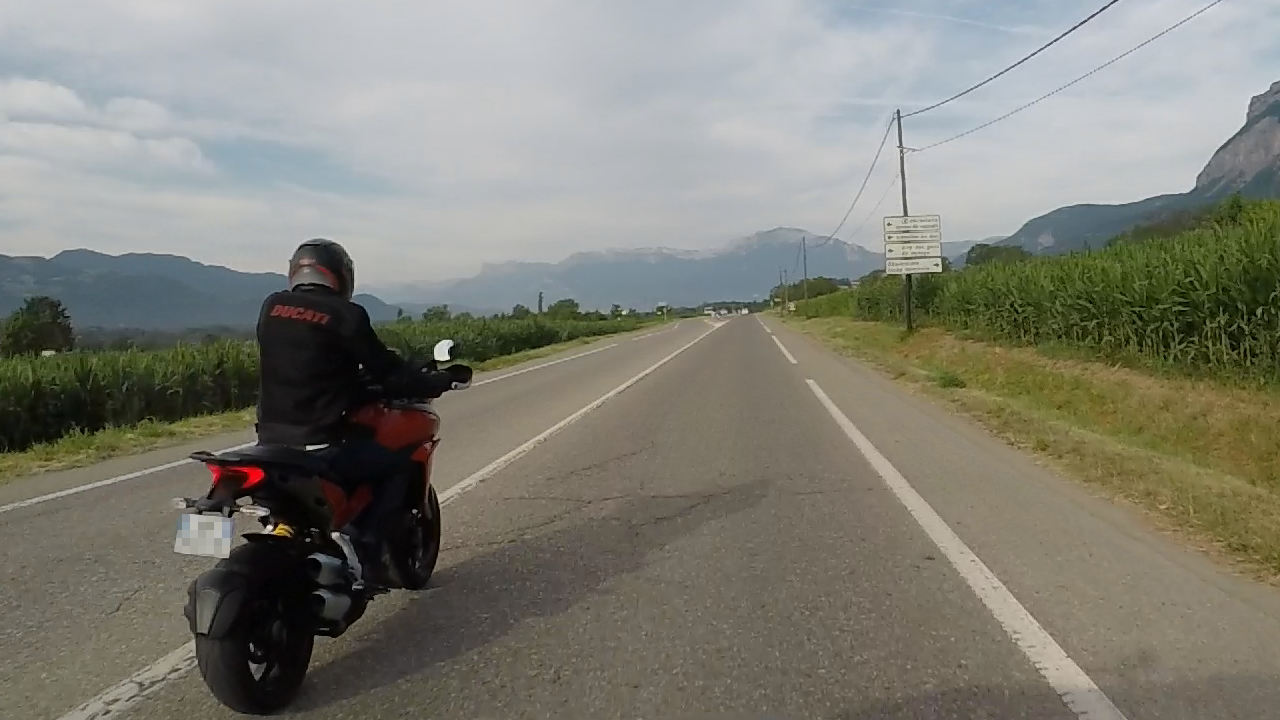} &
        \includegraphics[width=0.22\linewidth]{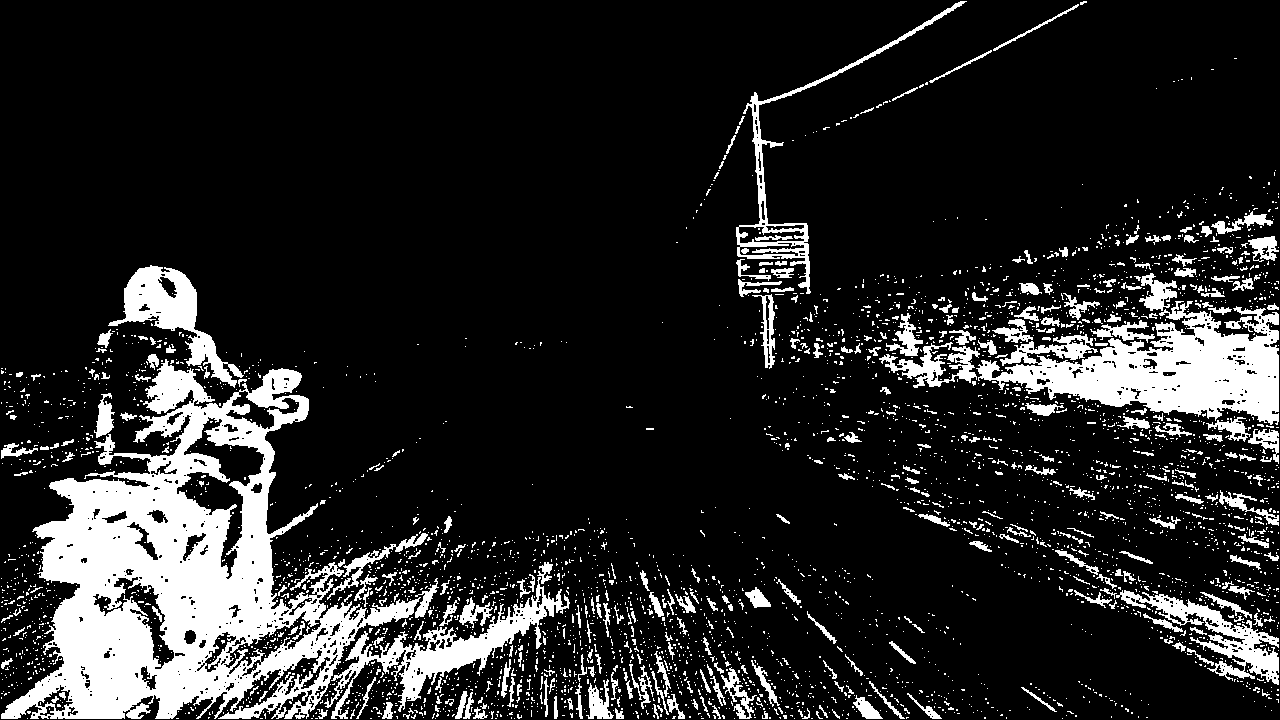} &
        \includegraphics[width=0.22\linewidth]{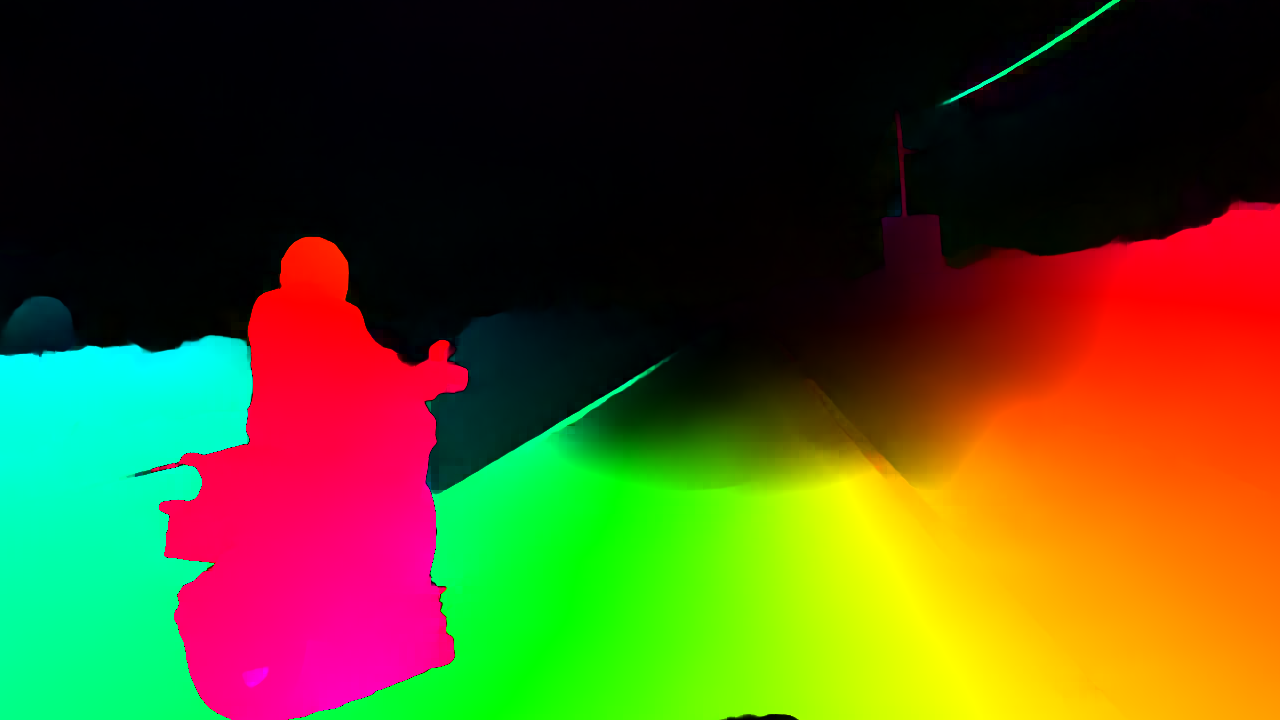} &
        \includegraphics[width=0.22\linewidth]{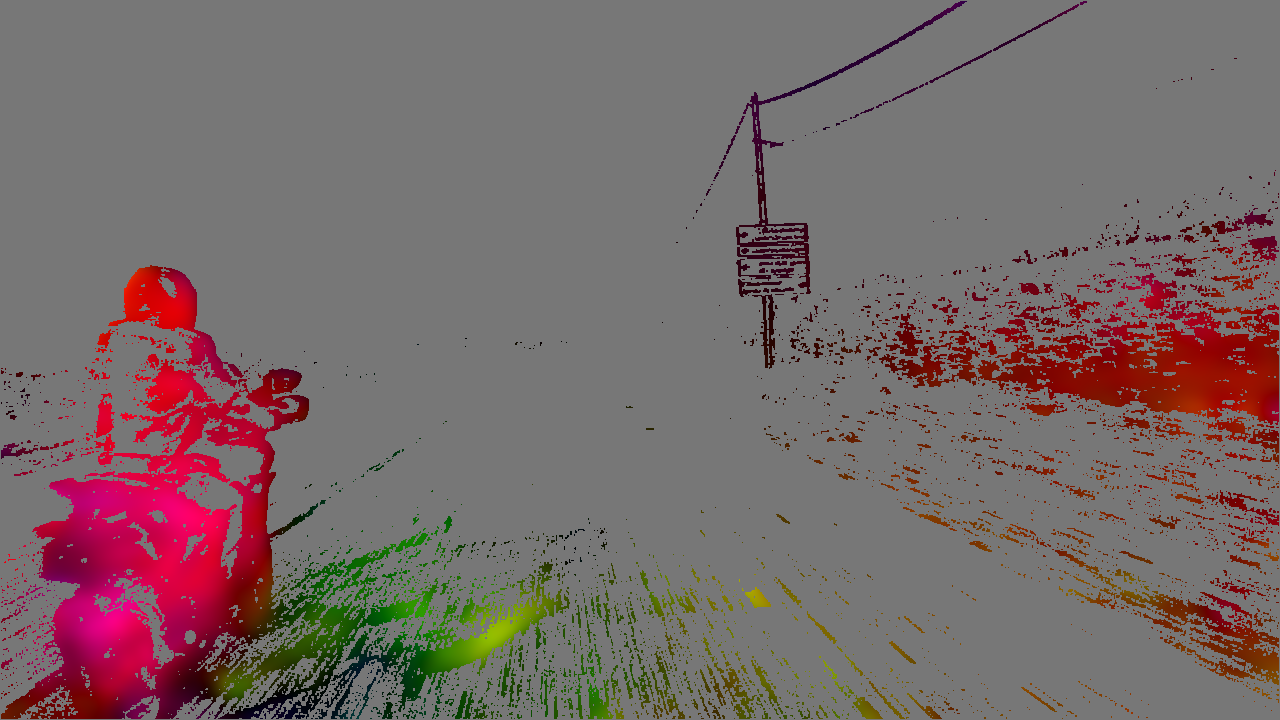} \\ % 276.7s
        \includegraphics[width=0.22\linewidth]{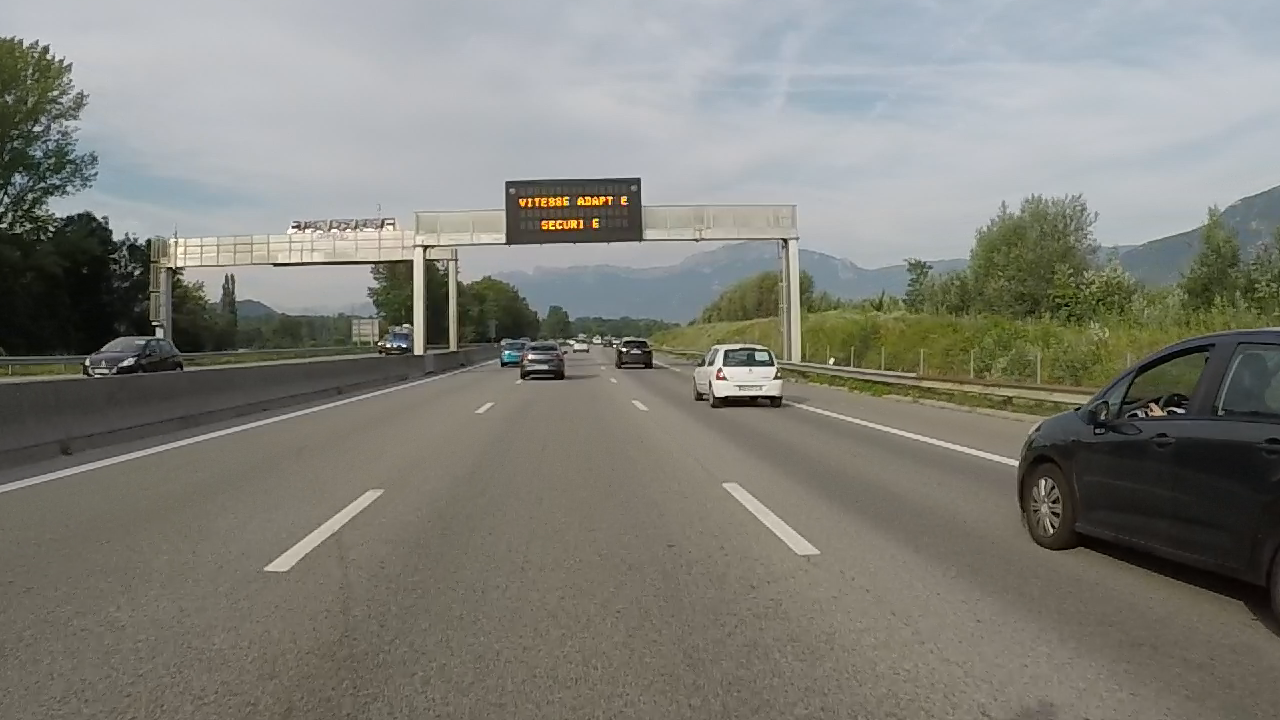} &
        \includegraphics[width=0.22\linewidth]{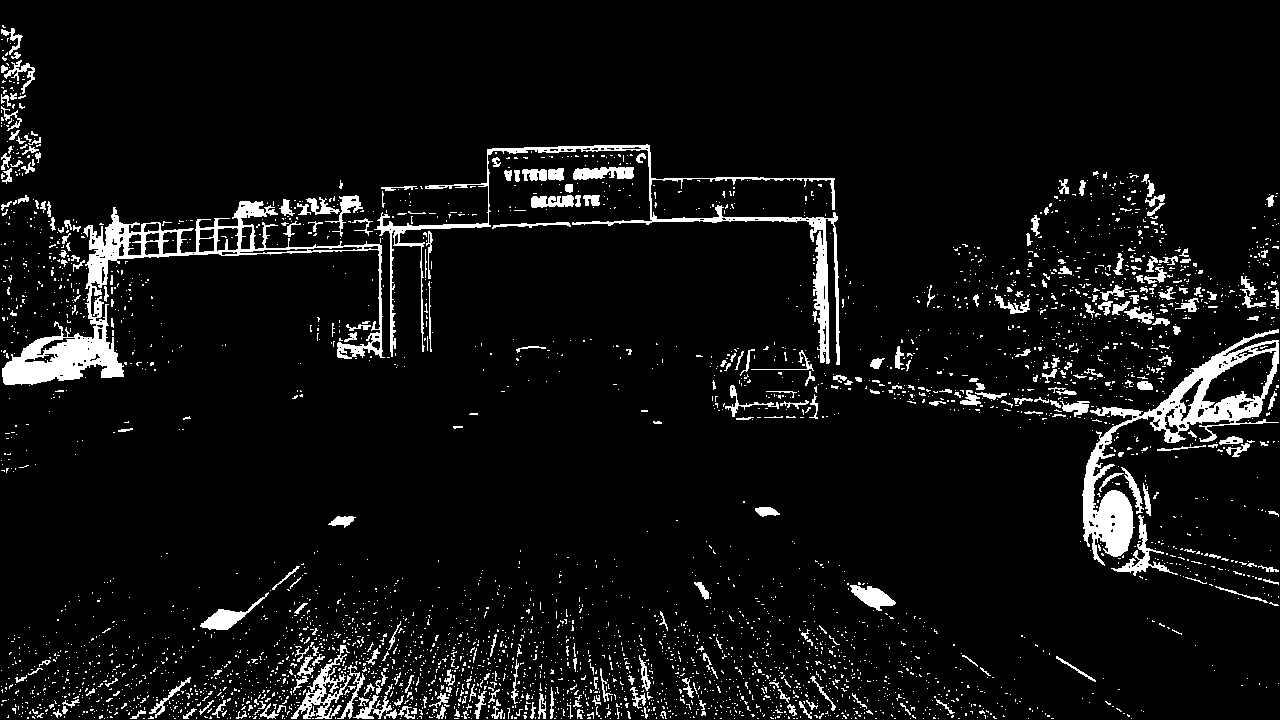} &
        \includegraphics[width=0.22\linewidth]{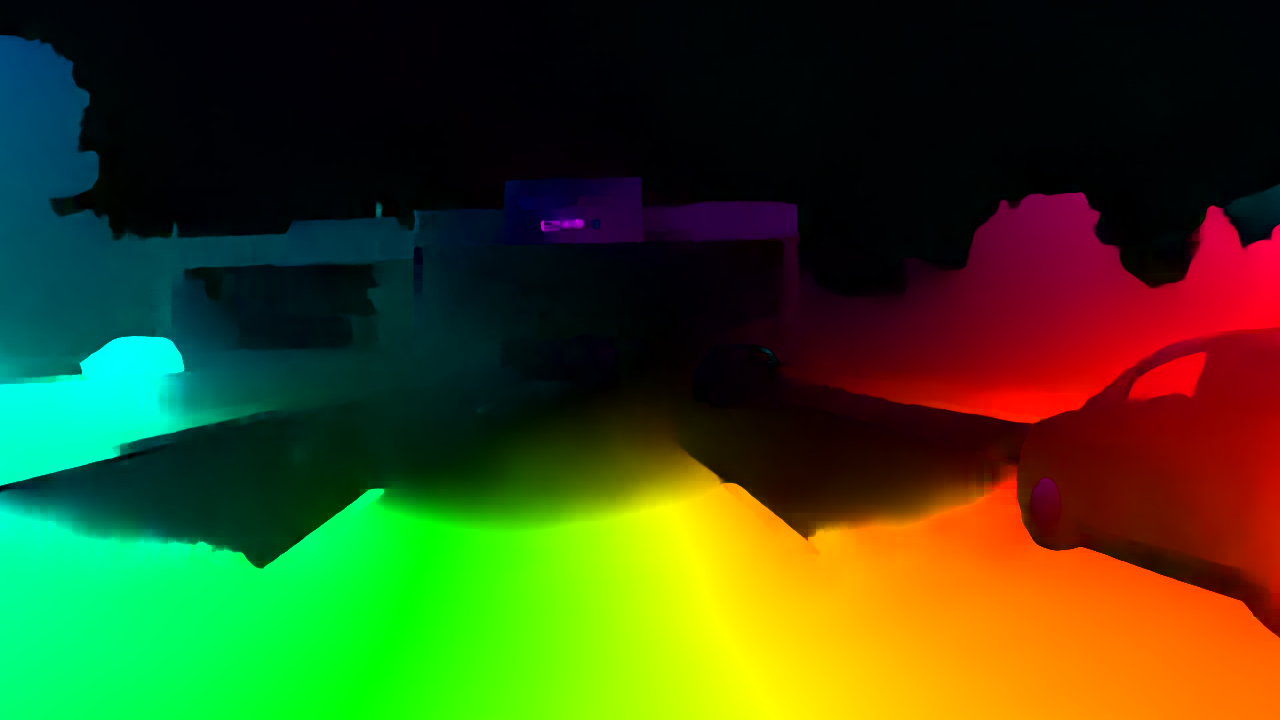} &
        \includegraphics[width=0.22\linewidth]{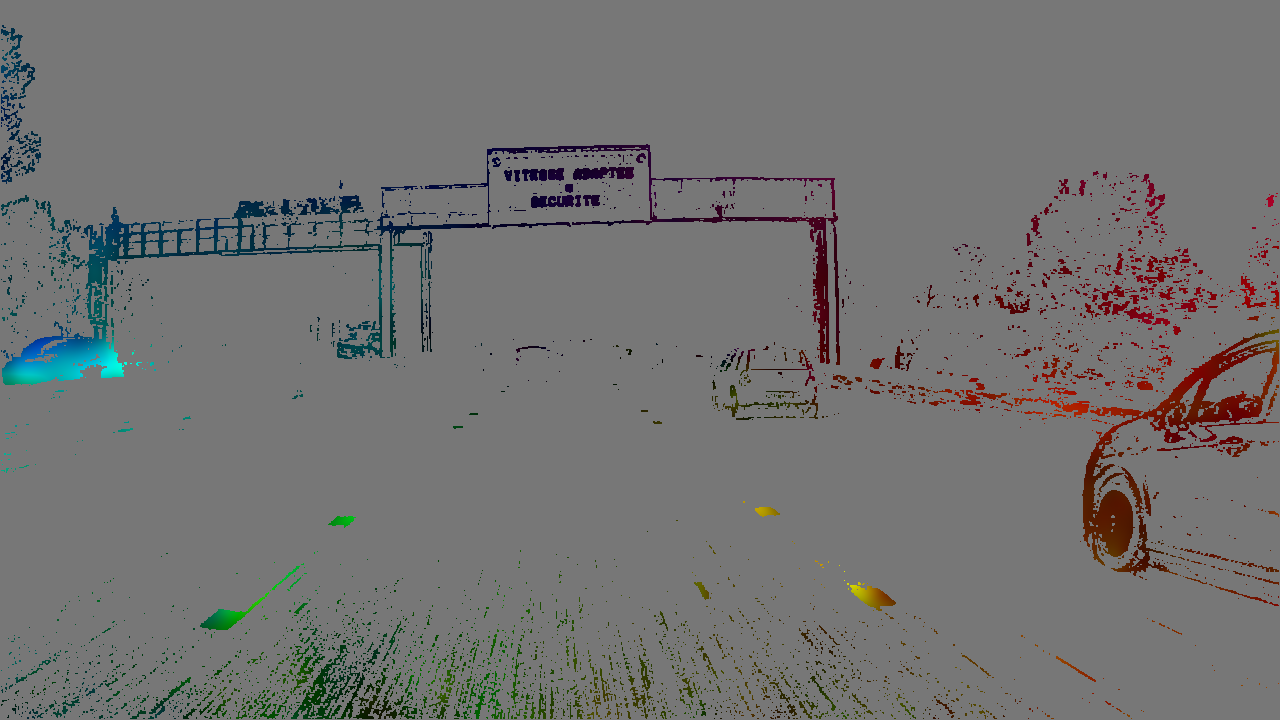} \\ % 509.5s
        \includegraphics[width=0.22\linewidth]{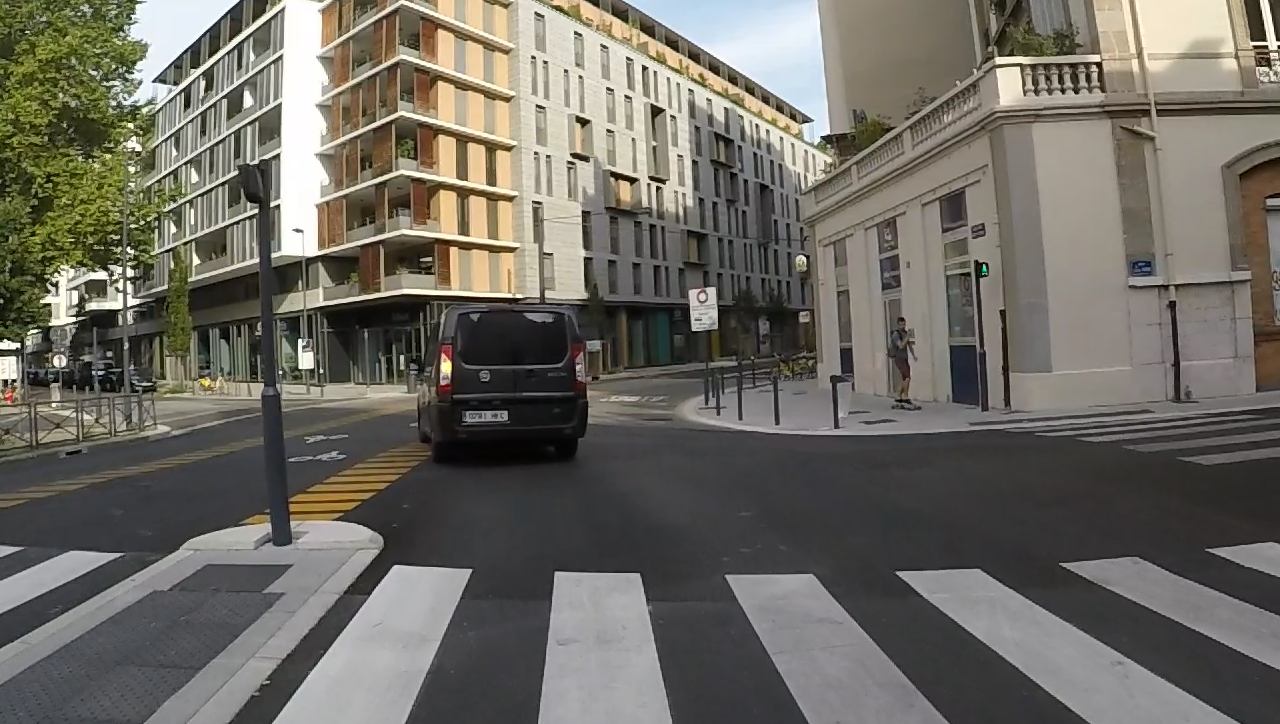} &
        \includegraphics[width=0.22\linewidth]{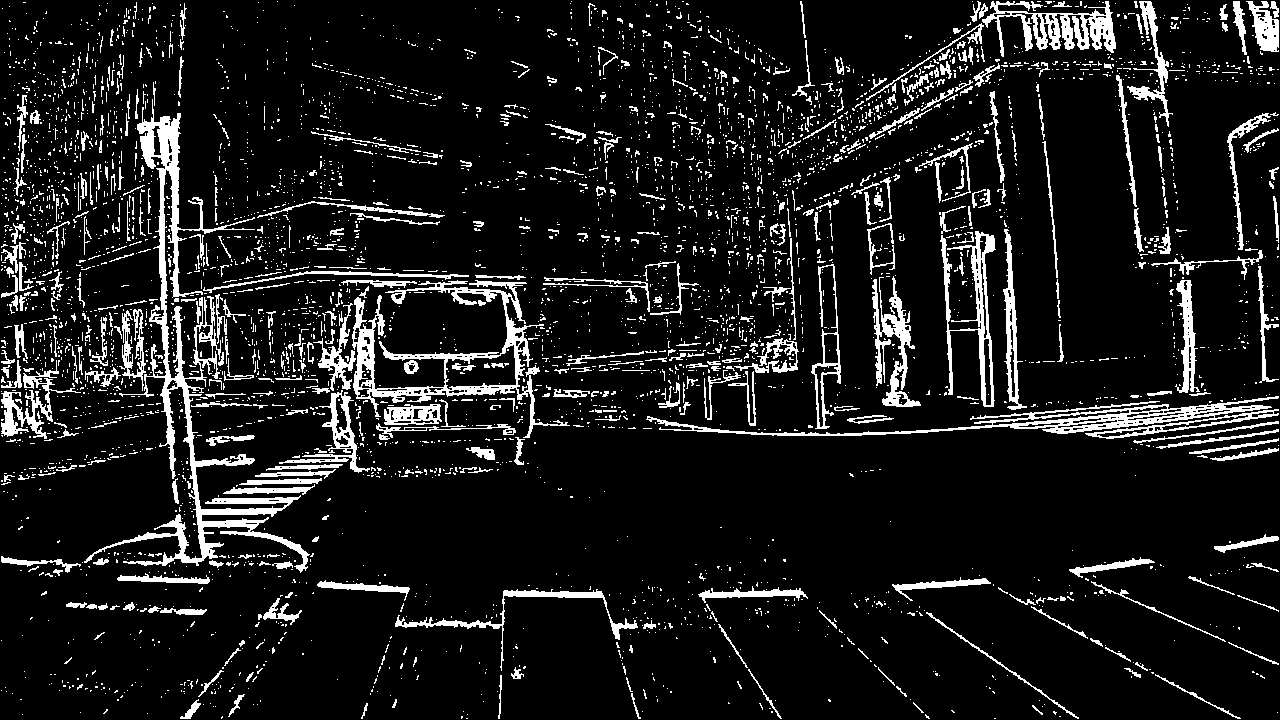} &
        \includegraphics[width=0.22\linewidth]{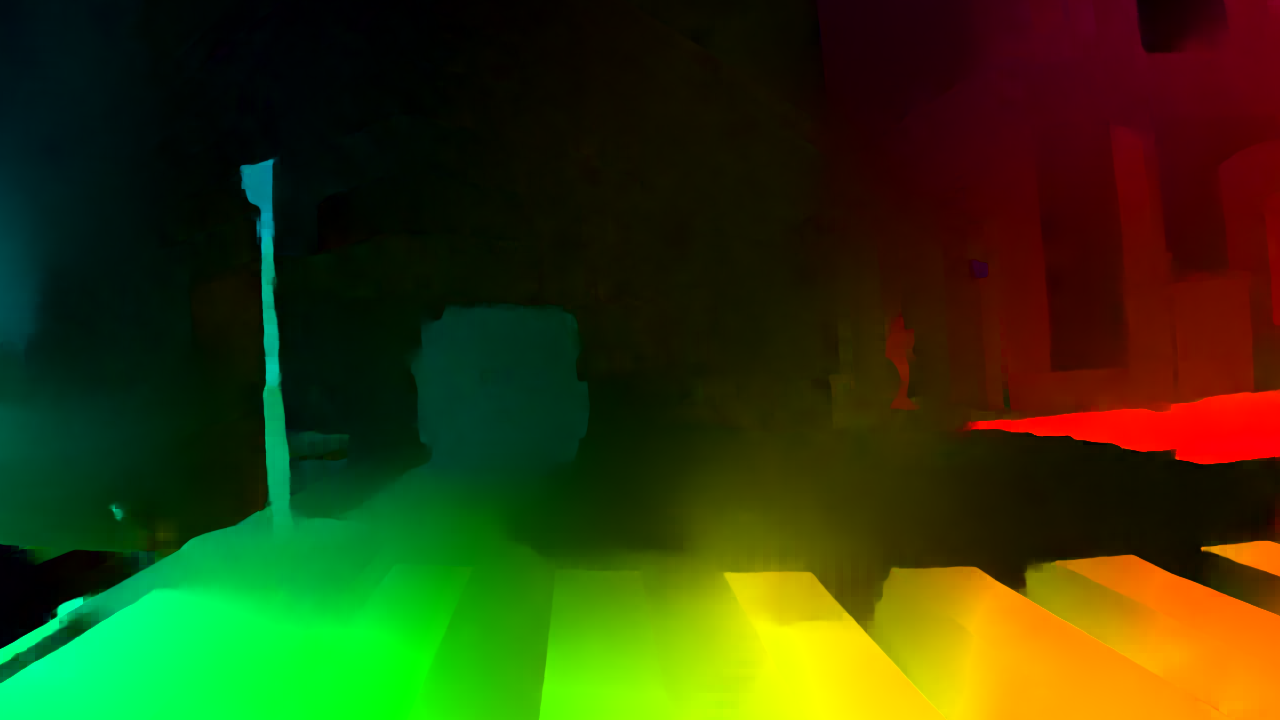} &
        \includegraphics[width=0.22\linewidth]{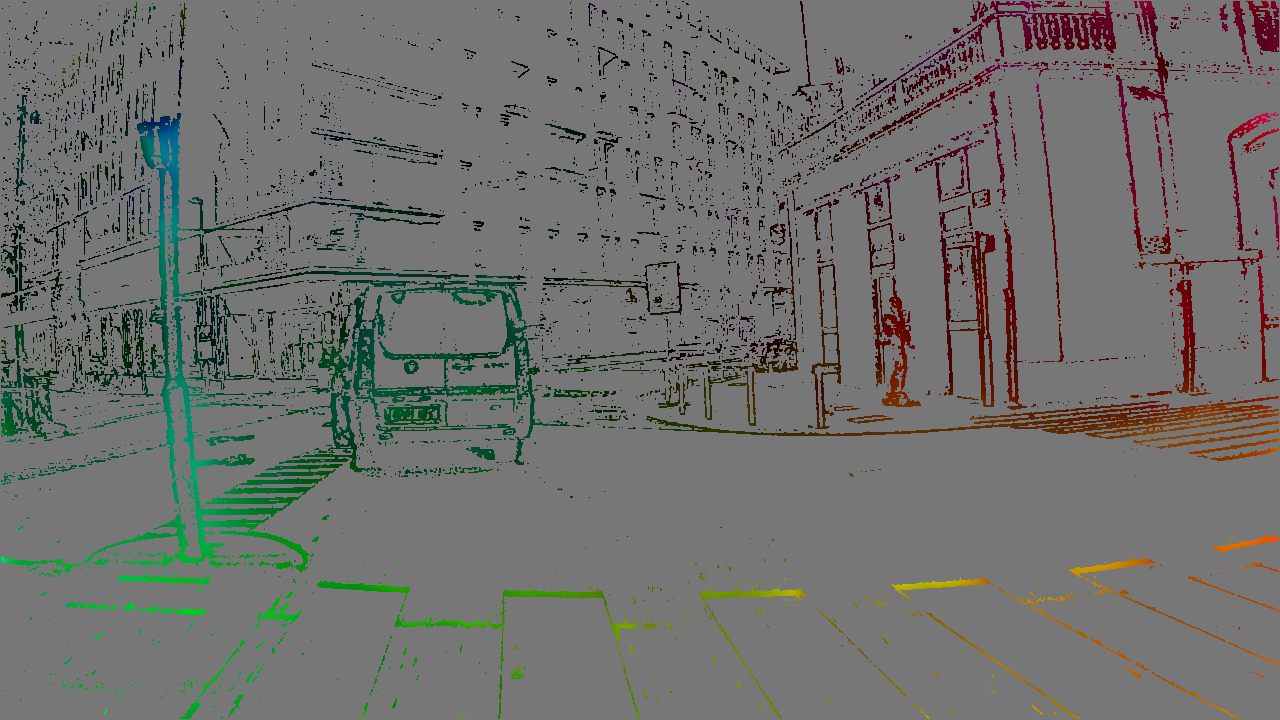} \\ % 1224.5s
    \end{tabular}
    \caption{Qualitative results on the 20-minute-long driving sequence. Extracts used, from top to bottom: a village street, a motorcycle overtaking, a highway, and an intersection. Best viewed in color.}\label{fig:results_hd}
\end{figure*}

We therefore present our FWL results on this dataset in Table~\ref{tab:fwl_1mp_results}, split between each recording day. Similarly to what was observed for low-resolution data, our method always yields FWL values greatly superior to \(1\), indicating an accurate optical flow. Compared to the ablation alternatives, it can also be observed that our final version displays the best global result, and the best results in all sequences but four (``Feb.\ 15'', ``Apr.\ 12'', ``Apr.\ 18'', and ``Jun.\ 26'', where it is the second best alternative), showing once again the importance of the denoising and of our inverse exponential distance surface.

\subsection{Complementary Evaluation on a 20-minute-long High-Resolution Driving Sequence}
While Prophesee's 1 Megapixel Automotive Detection dataset allows for a reproducible evaluation of EBOF methods with the FWL, it contains only event recordings, preventing the comparison with frame-based state-of-the-art methods. In order to complete our evaluation, we make use in this subsection of a twenty-minute-long driving sequence, containing both the output from high-definition frame-based and event-based cameras. This sequence presents a wide diversity of driving situations (urban/rural roads, highway, roundabouts, many other vehicles, pedestrians, \dots). It has been recorded by and graciously shared with us by Prophesee.

\begin{table}
    \centering
    \caption{FWL results on the 20-minute-long driving sequence}\label{tab:fwl_20_min_results}
    \resizebox{\linewidth}{!}{
        \begin{tabular}{c c c c c c}
        \toprule
        Village & Side Road & Highway & Suburban & Urban & \textit{Full sequence} \\
        (0'00 - 4'00) & (4'00 - 7'00) & (7'00 - 11'00) & (11'00 - 14'30) & (14'30 - 20'45) & (0'00 - 20'45) \\
        \midrule
        \midrule
        1.70 & 1.45 & 1.67 & 1.62 & 1.43 & 1.56 \\
        \bottomrule
        \end{tabular}
    }
\end{table}

The FWL results of this evaluation are presented in Table~\ref{tab:fwl_20_min_results}, split between each period of the sequence. An overall FWL result on the complete sequence is also presented. It can be observed that our FWL results are greatly satisfying, by remaining quite over the value of \(1\).

The main advantage of this sequence, however, lies in the possibility to compare our EBOF results to frame-based ones. For this prospect, we used RAFT~\cite{Teed2020RAFTRA} as frame-based reference, as it currently stands as one of the state-of-the-art optical flow methods. Due to the lack of calibration between the two sensors, only a qualitative evaluation can be presented. Therefore, several visual optical flow results are given in Fig.~\ref{fig:results_hd}. Rendering on the full sequence can be viewed in video format, at the link given at the beginning of this article. It can be seen that, similarly to when a low-resolution input is used, our optical flow remains visually very close to the reference.

\subsection{Evaluation on our High-Speed High-Definition Event-Based Indoor Dataset}\label{sec:highspeed_eval}

\begin{figure*}
    \centering
    \begin{tabular}{cccc}
        \includegraphics[width=0.22\linewidth]{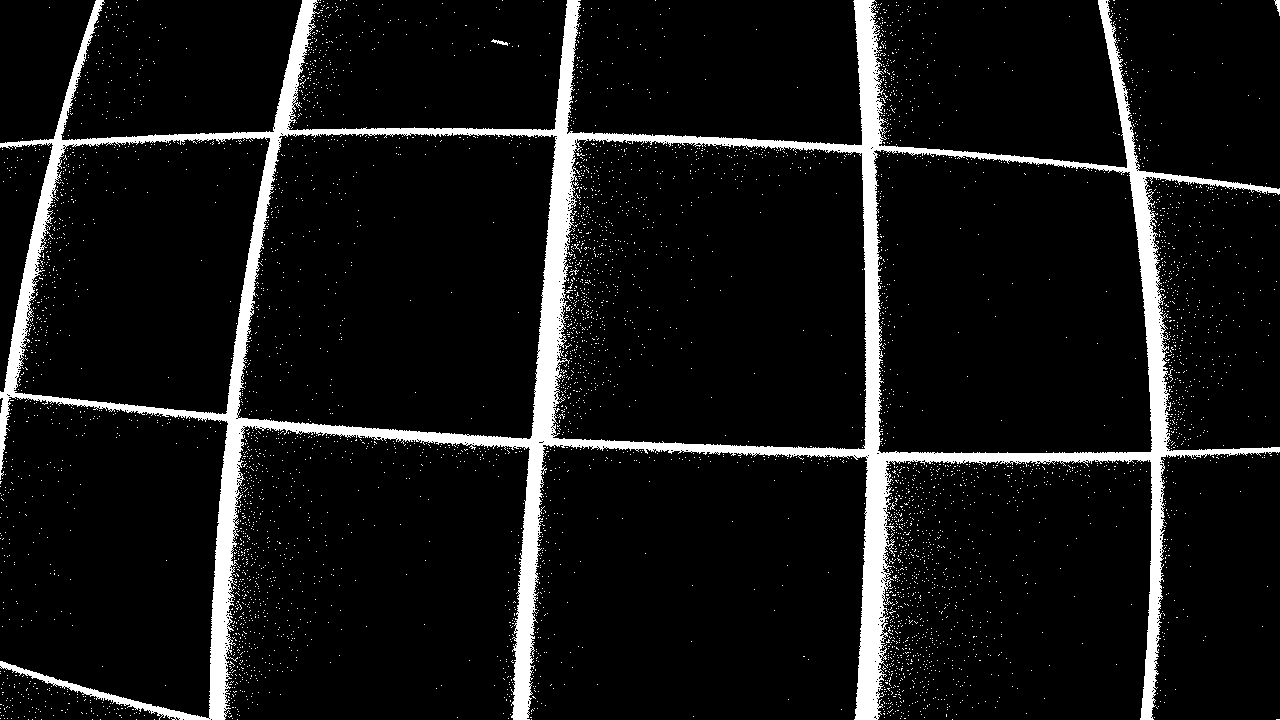} &
        \includegraphics[width=0.22\linewidth]{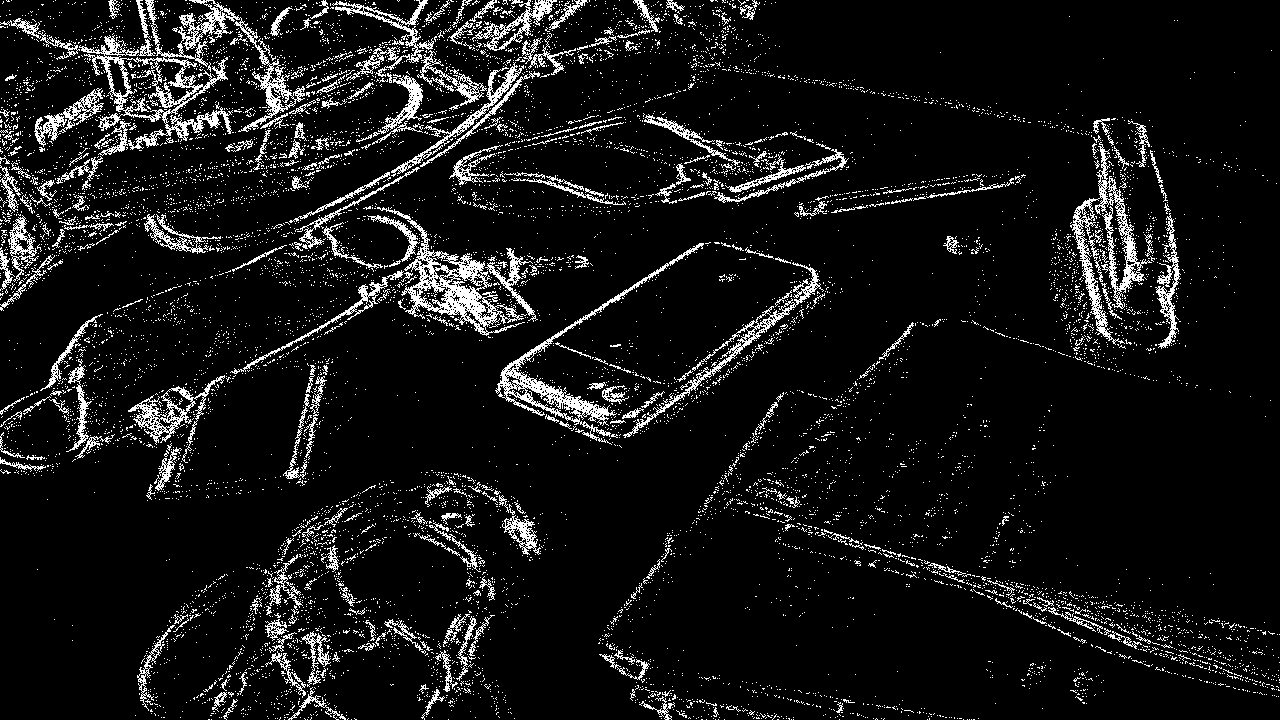} &
        \includegraphics[width=0.22\linewidth]{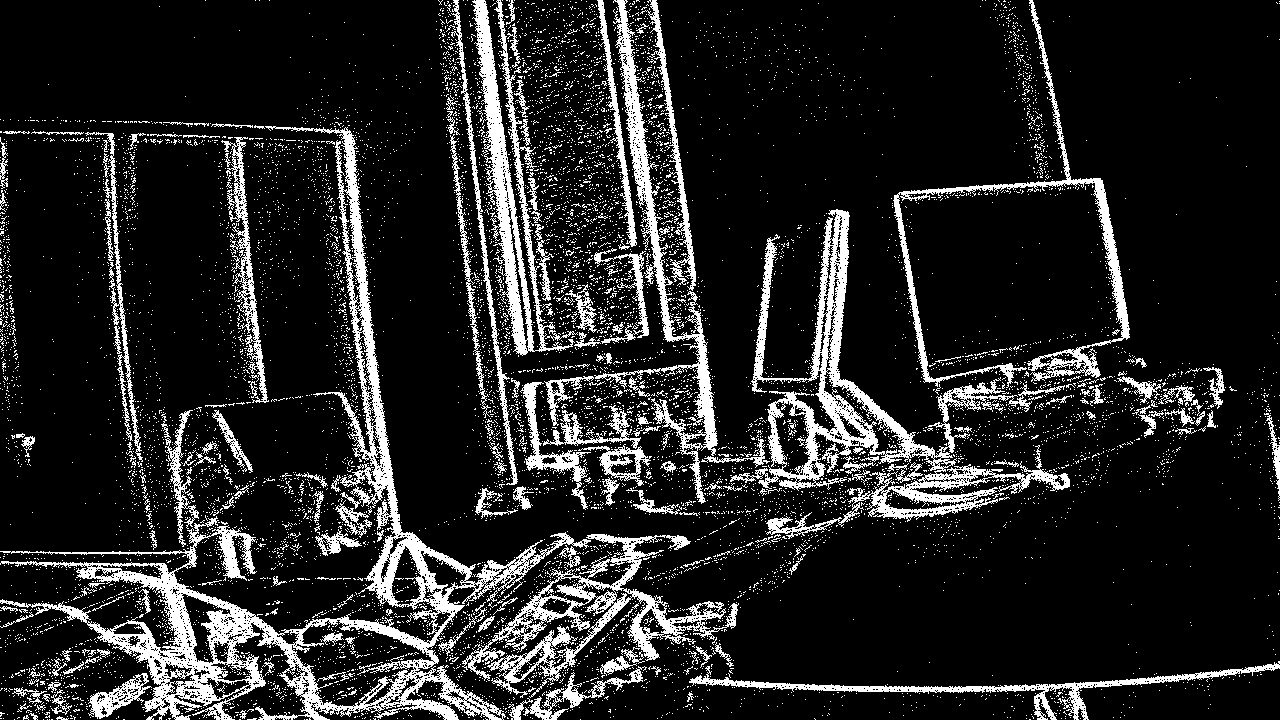} &
        \includegraphics[width=0.22\linewidth]{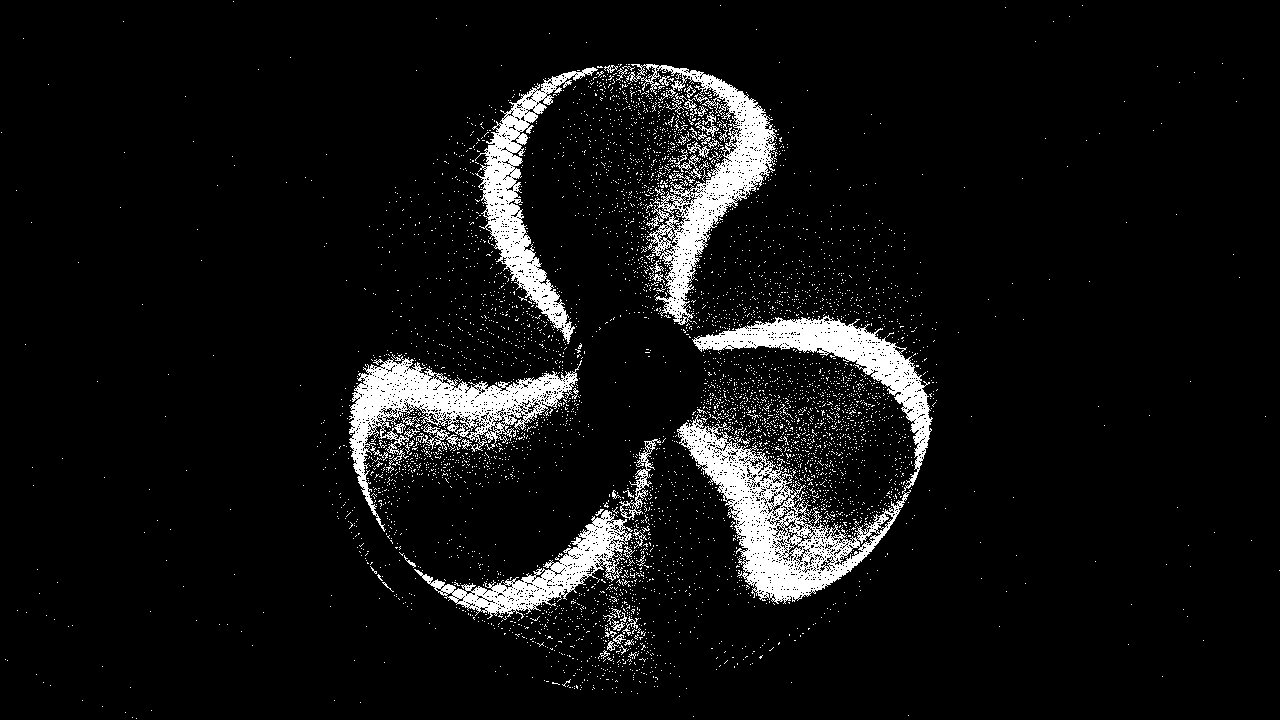} \\
    \end{tabular}
    \caption{Edge images of our high-speed high-definition event-based indoor dataset sequences. From left to right: ``Checkerboard'', ``Desk'', ``Office'', ``Fan''.}\label{fig:high_speed_hd_dataset}
\end{figure*}

For event-based driving sequences, most of our optical flow results are restricted to a few pixels, due to the low accumulation time of \(\Delta T = 15\text{ms}\) we used throughout this article, in accordance to movements speed. In order to show how our method is able to handle movements of higher magnitudes in various situations, we recorded a high-speed high-definition event-based dataset, using a Prophesee Gen4 camera (\(1280\times720\))~\cite{Finateu2020510A1}. This dataset is composed of four indoor sequences taken in office environment (namely, ``Checkerboard'', ``Desk'', ``Office'', and ``Fan''). The first three of them were recorded by manually shaking the camera, while for the last one, the camera was fixed in front of a high-speed fan. An illustration of these sequences is given in Fig.~\ref{fig:high_speed_hd_dataset}.

\begin{table}
    \centering
    \caption{FWL results on our high-speed high-definition event-based indoor dataset}\label{tab:fwl_our_hd_results}
    \begin{tabular}{c c c c c}
    \toprule
    Sequence & Checkerboard & Desk & Office & Fan \\
    \midrule
    \midrule
    \(\Delta T = 15\text{ms}\) & 1.49 & 1.71 & 1.74 & 1.05 \\
    \(\Delta T = 5\text{ms}\) & 1.75 & 1.66 & 1.84 & 1.53 \\
    \bottomrule
    \end{tabular}
\end{table}

\begin{figure}
    \centering
    \begin{tabular}{cc}
        \includegraphics[width=0.45\linewidth]{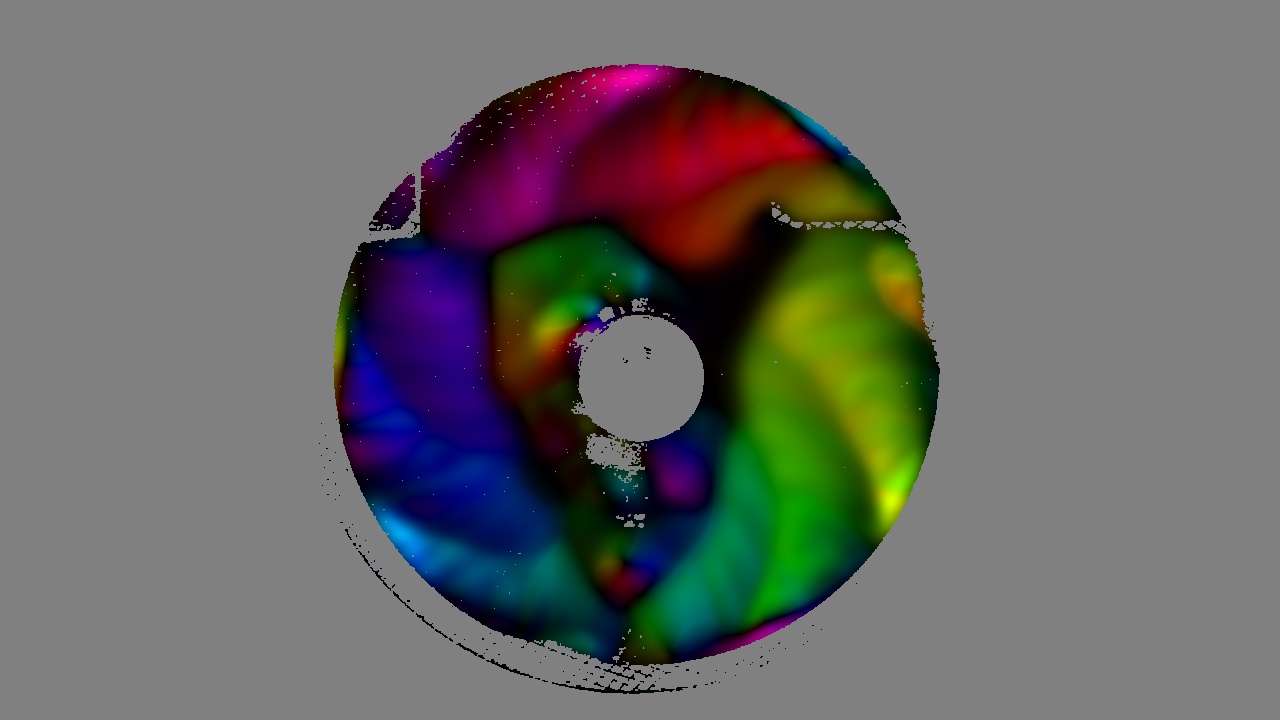} &
        \includegraphics[width=0.45\linewidth]{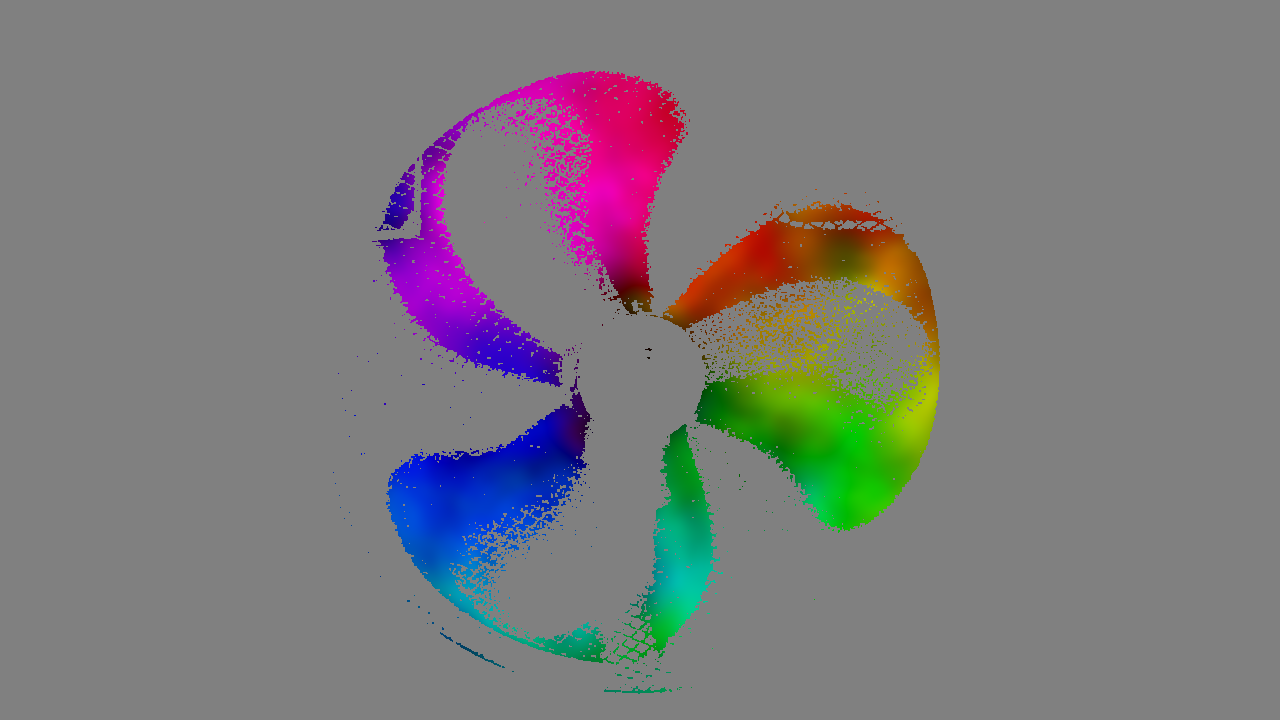} \\
    \end{tabular}
    \caption{Sample EBOF results for the ``Fan'' sequence of our high-speed dataset, with \(\Delta T = 15\text{ms}\) (left) and \(\Delta T = 5\text{ms}\) (right). Notice how the blades of the fan are merged together in the first case, while they appear clearly in the second one, leading to improved optical flow results.}\label{fig:high_speed_fan}
\end{figure}

This evaluation relies also on the FWL metric, to compare ourselves to the ``zero flow'' reference. The results are presented in Table~\ref{tab:fwl_our_hd_results}. It can be seen here that we always obtain a FWL greater than \(1\), underlining once again the accuracy of our optical flow results, even under larger apparent motions. However, apart for the ``Desk'' recording, all recordings display better FWL results when a lower accumulation time of \(\Delta T = 5\text{ms}\) is employed. This is due to the fact that, at such high motion speeds, the edge images become slightly too blurry to provide the best optical flow results when the accumulation time of \(\Delta T = 15\text{ms}\) is employed. Lowering accumulation time helps obtaining sharpest edge images, and in return, more accurate optical flow and FWL results. This remark is especially true for the ``Fan'' sequence, where the very high speed of the blades leads them to appear too blurry for a correct optical flow to be computed when \(\Delta T = 15\text{ms}\) is used; an illustration is given in Fig.~\ref{fig:high_speed_fan}.

\subsection{Sensitivity Analysis}\label{ssec:sensitivity}
\begin{figure}
    \centering
    \includegraphics[trim={0.1cm 0 0.45cm 0},clip, height=0.600\linewidth]{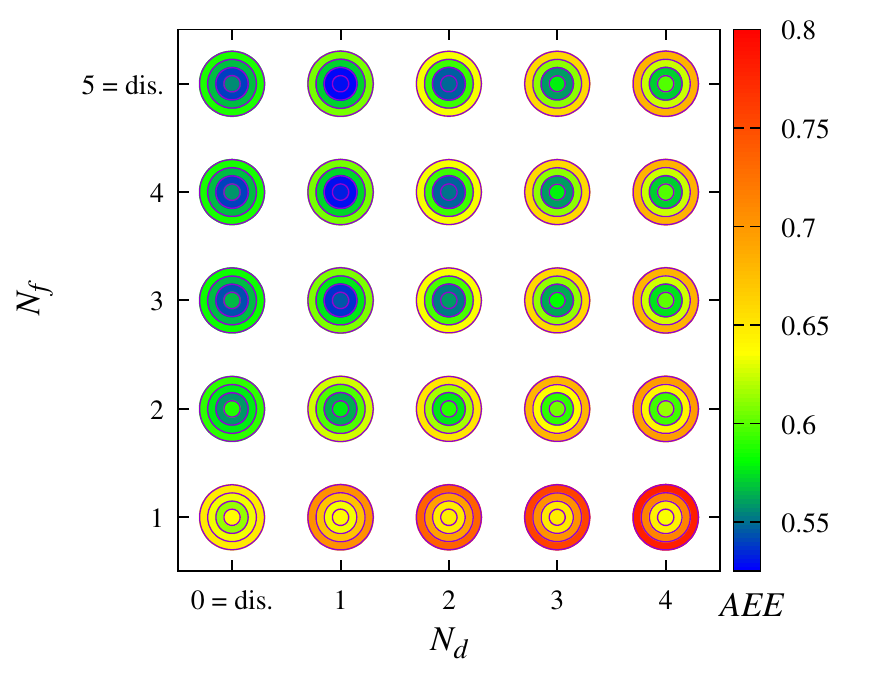} %&
    \hspace{2pt}
    \includegraphics[trim={4.05cm 0 2.2cm 0},clip, height=0.600\linewidth]{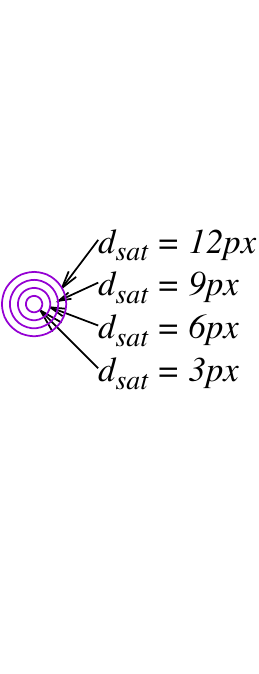} \\
    \caption{Sensitivity analysis of our EBOF method, using the AEE on the ``Outdoor day 1'' sequence from the MVSEC dataset, along the \(N_d\), \(N_f\), and \(d_\text{sat}\) parameters. Lowest AEE (blue) is the best.}\label{fig:sensitivity_analysis}
\end{figure}

In this subsection, we analyze the sensitivity of \(N_d\), \(N_f\) and \(d_\text{sat}\) parameters on our EBOF results. \(N_d\) and \(N_f\) define denoising and filling thresholds (see Section~\ref{ssec:denoisingfilling}), \(d_\text{sat}\) defines the saturation value for the computation of the inverse exponential distance surface (Section~\ref{sec:inv_exp_dist_surf}). For that purpose, we use the ``Outdoor day 1'' sequence from the MVSEC dataset with the AEE metric. We display, in Fig.~\ref{fig:sensitivity_analysis}, these results for \(N_d \in \{0=\text{``disabled''}, 1, 2, 3, 4\}\), \(N_f \in \{1, 2, 3, 4, 5=\text{``disabled''}\}\), and \(d_\text{sat} \in \{3, 6, 9, 12\}\) pixels.

\begin{table*}
    \centering
    \caption{Average execution times and standard deviation of each step of our method for low- and high-resolution inputs, in milliseconds}\label{tab:execution_times}
    \begin{tabular}{c c c c c c}
    \toprule
    Version & Edge image (after accumulation)\textsuperscript{*} & Denoising \& filling & Inverse exponential distance transform & Optical flow\textsuperscript{\textdagger} & Total \\
    \midrule
    \midrule
    \multicolumn{6}{c}{\textbf{Low-resolution (\(\mathbf{346\times260}\))}} \\
    CPU-only & 0.15\rpm{0.03} & 0.77\rpm{0.16} & 5.07\rpm{2.07} & \textit{3.59\rpm{0.09}} & 9.57\rpm{2.13} \\
    CPU \& GPU & 0.18\rpm{0.04} & 0.50\rpm{0.04} & 1.37\rpm{0.09} & 3.76\rpm{0.16} & 5.82\rpm{0.23} \\
    \midrule
    \midrule
    \multicolumn{6}{c}{\textbf{High-resolution (\(\mathbf{1280 \times 720}\))}} \\
    CPU-only & 0.54\rpm{0.06} & 3.55\rpm{0.30} & 7.43\rpm{0.51} & \textit{12.21\rpm{0.16}} & 23.73\rpm{0.69} \\
    CPU \& GPU & 0.55\rpm{0.07} & 0.69\rpm{0.14} & 3.75\rpm{0.46} & 11.89\rpm{0.76} & 16.88\rpm{1.30} \\
    \bottomrule
    \multicolumn{6}{l}{\textsuperscript{*}This module uses only the CPU, hence the similar results between the ``CPU-only'' and ``CPU \& GPU'' lines for this column.} \\
    \multicolumn{6}{p{485pt}}{\textsuperscript{\textdagger}The optical flow library provided by Adarve \textit{et al.}~\cite{Adarve2016AFF} does not contain a CPU-only version. The CPU-only experiments therefore use the GPU for optical flow computation, hence the similar results between the ``CPU-only'' and ``CPU \& GPU'' lines for this column.}
    \end{tabular}
\end{table*}

From this plot, it can first be noted that the denoising should not be too strong, that is, \(N_d \leq 2\). In the same time, the filling threshold should be \(N_f \geq 3\). Overall best \(d_{\text{sat}}\) value is incontestably \(d_{\text{sat}} = 6\text{px}\), and it is the most sensitive parameter. The best set of parameters is \(\{N_d = 1, N_f = 4, d_\text{sat} = 6\}\), with the corresponding minimal AEE = 0.53px. The worst set of parameters is \(\{N_d = 4, N_f = 1, d_\text{sat} = 12\}\), with the corresponding AEE = 0.78px, increasing by 47\% compared to the best parameters. This shows the importance of the parameters choice to guarantee adequate optical flow results.

\subsection{Real-Time Compliance}
To finally show the real-time compliance of our approach for both low- and high-resolution input data, we used respectively the ``Indoor flying 1'' sequence from the MVSEC dataset and the \texttt{moorea\_2019-01-30\_000\_td\_671500000\_\linebreak731500000\_td} test sequence from Prophesee's 1 Megapixel Automotive Detection dataset. The execution time for each block was not evaluated separately, but simultaneously, as the full pipeline should be able to run on a single computer. In order to also underline the benefits brought by the use of the GPU, we ran the evaluation on both the CPU-only and the CPU+GPU versions of our code.

These results are presented in Table~\ref{tab:execution_times}. From them, it can be seen on one hand that the GPU-aided version can achieve real-time performances for both low- and high-resolution input data, if their accumulation times are set to at least 4ms and 13ms respectively (the limiting factor being the optical flow module). The values for \(\Delta T\) used throughout this article, therefore, respect the real-time constraint. The CPU-only version, on the other hand, can achieve real-time performances for both low- and high-resolution data when the accumulation time is of at least 8ms and 13ms respectively (the limiting factor being here the distance transform and optical flow module respectively). As noted in Table~\ref{tab:execution_times}, however, the optical flow can only be computed on GPU at this time, and transferring this computation to the CPU would likely make these execution times greatly increase.

From the results of Table~\ref{tab:execution_times} and from the previous conclusions, using our architecture, the best performances that can be reached for a low-resolution input is therefore a 250Hz flow with a latency of 10ms (for the minimal \(\Delta T = 4\)ms of accumulation time), while, for a high-resolution input, a 77Hz flow with a latency of 30ms can be achieved (for the minimal \(\Delta T = 13\)ms of accumulation time).

As a fast algorithm, FireFlowNet~\cite{ParedesValls2021BackTE} displays a theoretical inference frequency up to 262Hz for a low-resolution (\(346\times260\)) input, similar to ours. For a higher definition (\(1280\times720\)), however, our approach achieves frame rates 2- to 3-times better than them, as their method can only reach 29Hz. The GPU they use ranks similarly to ours in popular benchmarks. In addition, we ran EV-FlowNet~\cite{Zhu2018EVFlowNetSO} on low- (\(346\times260\)) and on high-resolution (\(1280\times720\)) data. While EV-FlowNet inference achieved a 125Hz flow on low-definition data, it showed its limits on high-resolution input with a 12.5Hz flow output. These frequencies consider only inference process, full latency is unknown and should include events accumulation time and specific dense representation creation.

To compare with a state-of-the-art frame-based optical flow algorithm, we measured a 12.5Hz flow on low-resolution input, and a 2Hz one on high-definition images with RAFT~\cite{Teed2020RAFTRA}.

\section{Conclusion}
In this article, a complete pipeline for real-time computation of event-based optical flow (EBOF) from both low- and high-resolution event cameras has been proposed. It includes optimized algorithmic choices as well as a novel inverse exponential distance surface representation. Several evaluations have been conducted to show the relevance of our contributions. Resulting accuracies surpass or are close to the non-real-time state of the art for low-definition recordings, as well as on novel high-definition sequences. Frame rates of respectively 250Hz and 77Hz for resolutions of \(346\times260\) and \(1280\times720\) were also achieved, making it, to the best of our knowledge, the most accurate EBOF method for low- and especially high-resolution event cameras that could be deployed in the wild.

Deep learning dominates EBOF estimation when looking for accuracy at the cost of real-time ability. The EV-FlowNet architecture is a perfect example of this trend, as shown in Table~\ref{tab:mvsec_results}. On the contrary, when real-time running is needed, this is currently our proposition, which is not a neural network, that provided the best results. As our method is not based on machine learning, it is independent from a training process involving specific datasets and loss functions. In other words, the results are expected to be similar to the ones presented here for all types of scenes. This can be seen as an advantage, compared for example to EV-FlowNet which shows very different results according to the way it has been trained.

In hindsight, improvements could be brought to the current method, especially regarding the event accumulation process. Relying on a predetermined fixed accumulation time \(\Delta T\), which highly depends on the appearance and dynamics of the visual scene, can indeed lead to instabilities in the appearance of the edge images if not chosen carefully. Introducing an adaptive method to dynamically determine the correct accumulation time to use, as proposed in~\cite{Liu2018AdaptiveTB} for instance, could allow for obtaining clear edge images independently from the scene evolution, but would also certainly make the real-time constraint harder to achieve. Our architecture could also benefit from a more optimized implementation on specialized hardware (FPGA for instance, eliminating the need for an energy-intensive GPU), which could also allow for even higher frame rates by lowering the minimum accumulation times. Applying our EBOF to complex automotive-related applications, such as proposed in~\cite{Jung2020ConstrainedFF}, could also be addressed by future work, for instance for improving the detection of obstacles appearing in the close neighbourhood of the vehicle.

\section*{Acknowledgment}
The authors thank Renault for lending their Prophesee Gen 4 camera, Prophesee for giving us access to their twenty-minute-long driving sequence, and in particular Davide Migliore for providing thoughtful insight on their high-definition event sensor.

\bibliography{bibliography}

\begin{IEEEbiography}[{\includegraphics[width=1in,height=1.25in,clip,keepaspectratio]{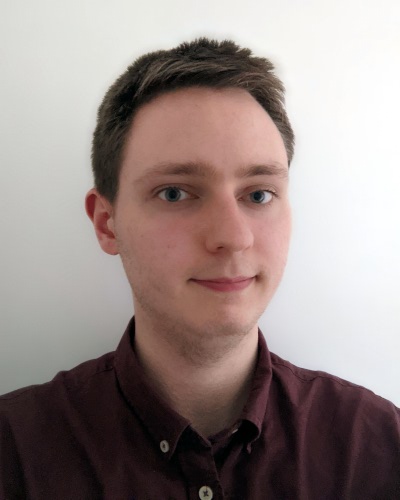}}]{Vincent Brebion}
was born in Blendecques, France, in 1997. He received the engineering degree in computer science and the master's degree in complex systems engineering from the Universit\'{e} de technologie de Compi\`{e}gne (UTC), France, in 2020. He is currently pursuing the Ph.D.\ degree in computer vision at the Heudiasyc Lab., UTC, France, and is currently a member of the UTC/CNRS/Renault SIVALab joint laboratory. His research interests include event-based cameras, multi-sensor fusion, and computer vision for autonomous vehicles.
\end{IEEEbiography}

\begin{IEEEbiography}[{\includegraphics[width=1in,height=1.25in,clip,keepaspectratio]{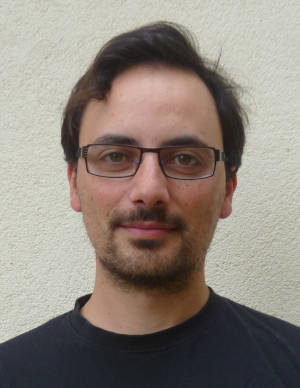}}]{Julien Moreau}
received the Ph.D. degree in computer vision from the Universit\'{e} de Technologie de Belfort-Montb\'{e}liard (UTBM), France, in 2016.
He accomplished postdoctoral positions in The French Institute of Science and Technology for Transport Development and Networks (IFSTTAR), in Lille, France, and in the Institute of Information and Communication Technologies Electronics and Applied Mathematics (ICTEAM), at Universit\'{e} Catholique de Louvain, Louvain-la-Neuve, Belgium.
Since 2019, he is an associate professor in the Computer Science department of Universit\'{e} de technologie de Compi\`{e}gne (UTC), France, and is carrying out his research in Heudiasyc UMR 7253, a joint UTC-CNRS research laboratory.
From that, he is also a member of SIVALab, a joint laboratory between Renault, UTC and CNRS.
His research interests cover stereovision, unconventional cameras, calibration and machine learning applied to perception and localization for mobile robotics.
\end{IEEEbiography}

\begin{IEEEbiography}[{\includegraphics[width=1in,height=1.25in,clip,keepaspectratio]{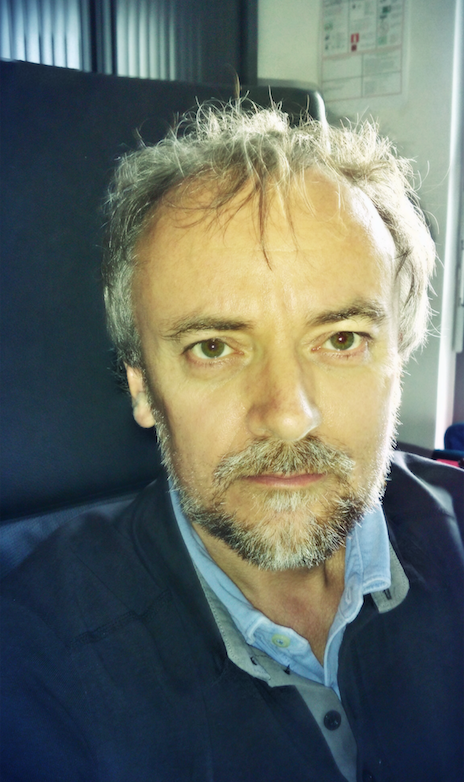}}]{Franck Davoine}
received his Ph.D. in 1995 from Grenoble INP - Universit\'{e} Grenoble Alpes in France.
He was appointed at Universit\'{e} de technologie de Compi\`{e}gne (UTC), Heudiasyc Lab., France, in 1997 as an Associate professor and in 2002 as a Researcher at CNRS.\ From 2007 to 2014, he was on leave at LIAMA Sino-European Lab.\ in Beijing, P.R. China, as PI of a project with CNRS and Peking University on Multi-sensor based perception and reasoning for intelligent vehicles. In 2015, he was back in Compiegne, PI of a challenge-team within the Laboratory of Excellence of UTC focusing on Collaborative vehicle perception and urban scene understanding for autonomous driving, and member of the CNRS/UTC/Renault SIVALab joint laboratory specializing in localization and perception systems for autonomous vehicles. 
\end{IEEEbiography}


%\vfill


\end{document}